\documentclass{article}

\usepackage{microtype}
\usepackage{graphicx}
\usepackage{booktabs} 
\usepackage{subcaption}
\usepackage{diagbox}

\usepackage{microtype}
\usepackage{graphicx}
\usepackage{hyperref}       
\usepackage{booktabs} 
\usepackage{multicol}
\usepackage{multirow, varwidth}
\usepackage{amssymb}
\usepackage{amsmath}
\usepackage{amsthm}
\usepackage{algorithm} 
\usepackage{algorithmic} 
\usepackage[dvipsnames]{xcolor}
\usepackage{colortbl}
\usepackage{natbib}
\usepackage{subcaption}
\usepackage{wrapfig}
\usepackage{xspace} 
\usepackage{enumitem}
\usepackage{tikz}
\usetikzlibrary{tikzmark,calc}
\usepackage{soul}

\usepackage{hyperref}

\usepackage[accepted]{icml2025}

\usepackage{amsmath}
\usepackage{amssymb}
\usepackage{mathtools}
\usepackage{amsthm}

\usepackage[capitalize,noabbrev]{cleveref}

\theoremstyle{plain}
\newtheorem{theorem}{Theorem}[section]

\newtheorem{lemma}[theorem]{Lemma}

\theoremstyle{definition}
\newtheorem{definition}[theorem]{Definition}

\theoremstyle{remark}

\newcommand{\var}{\textup{var}}

\usepackage[textsize=tiny]{todonotes}

\icmltitlerunning{Tilted Sharpness-Aware Minimization}

\begin{document}

\twocolumn[
\icmltitle{Tilted Sharpness-Aware Minimization}




\begin{icmlauthorlist}
\icmlauthor{Tian Li}{a}
\icmlauthor{Tianyi Zhou}{b}
\icmlauthor{Jeffrey A. Bilmes}{c}
\end{icmlauthorlist}

\icmlaffiliation{a}{University of Chicago}
\icmlaffiliation{b}{University of Maryland, College Park}
\icmlaffiliation{c}{University of Washington}

\icmlcorrespondingauthor{Tian Li}{litian@uchicago.edu}

\icmlkeywords{Sharpness-Aware Minimization, Generalization, Exponential Tilting}

\vskip 0.3in
]



\printAffiliationsAndNotice{}  

\begin{abstract}
Sharpness-Aware Minimization (SAM) has been demonstrated to improve the generalization performance of overparameterized models by seeking flat minima on the loss landscape through optimizing model parameters that incur the largest loss within a neighborhood. Nevertheless, such min-max formulations are computationally challenging especially when the problem is highly non-convex. Additionally, focusing only on the worst-case local solution while ignoring potentially many other local solutions may be suboptimal when searching for flat minima. In this work, we propose Tilted SAM (TSAM), a smoothed generalization of SAM inspired by exponential tilting that effectively assigns higher priority to local solutions that incur larger losses. TSAM is parameterized by a tilt hyperparameter $t$ and reduces to SAM as $t$ approaches infinity. We show that TSAM is smoother than SAM and thus easier to optimize, and it explicitly favors flatter minima. We develop algorithms motivated by the discretization of Hamiltonian dynamics to solve TSAM. Empirically, TSAM arrives at flatter local minima and results in superior test performance than the baselines of SAM and ERM across a range of image and text tasks.
\end{abstract}

\section{Introduction}
Empirical risk minimization (ERM) is a classic framework for machine learning that optimizes for the average performance of the observed samples. For $n$ training samples $\{x_i\}_{i \in [n]}$ (which may also contain label information), 
 model parameters $\theta \in \mathbb{R}^d$, and a loss function $l(\cdot)$, let ERM be defined as 
\begin{align}
    \min_{\theta} L(\theta) :=  \frac{1}{n} \sum_{i \in [n]} l(x_i;\theta). \label{obj:erm}
\end{align}
In overparameterized models, however, minimizing ERM may arrive at a bad local minimum. To address this, one line of work focuses on minimizing the sharpness of final solutions, ensuring that the losses of parameters around local minima are uniformly small. One popular formulation is sharpness-aware minimization (SAM), that optimizes over the worst-case loss over perturbed parameters~\cite{foret2020sharpness}. For a perturbing region $\|\epsilon\|\leq\rho$ where $\epsilon \in \mathbb{R}^d$, the canonical SAM  objective is defined as 
\begin{align}
    \min_{\theta} L^s(\theta) := \max_{\|\epsilon\| \leq \rho} L(\theta + \epsilon). \label{obj:sam}
\end{align}
Typically, SAM is optimized by alternating between running gradient ascent  (to find the max loss) and gradient descent steps (to minimize the max loss) on model parameters~\citep[e.g.,][]{foret2020sharpness,andriushchenko2022towards}. However, it is difficult for such algorithms  (and its variants) that rely on one or few steps of gradient ascent to find the exact perturbation $\epsilon$ that incurs the true max loss, as the loss landscape can be highly non-convex and potentially non-smooth. Despite recent advancements on approximately solving the SAM objective~\citep[e.g.,][]{liu2022random,liu2022towards,zhuang2022surrogate}, the min-max formulation itself overlooks many neighborhood regions that may also result in large losses, leaving some loss surface near local minima sharp.
For instance, we have computed the average loss across the neighborhoods of SAM solutions, and find that it is still higher than the ones obtained by our approach (Section~\ref{sec:experiments:sharpness}).

To this end, we propose a generalized and smoothed variant of SAM inspired by exponential tilting and its widespread usage in probability and statistics. In optimization literature, it has also been used as an efficient min-max smoothing operator~\citep{kort1972new}. Tilted SAM (TSAM), parameterized by a tilt scalar $t \geq 0$, is defined as 
\begin{align}
    \min_{\theta} L^t(\theta) &:= \frac{1}{t} \log \left(\int e^{tL(\theta+\epsilon)} d \mu(\epsilon) \right) \nonumber \\ &= \frac{1}{t} \log \left(\mathbb{E}_{\mu(\epsilon)} \left[e^{t L(\theta+\epsilon)}\right]\right), \label{obj:tsam}
\end{align}
where $L(\theta+\epsilon)$ is defined in Eq.~(\ref{obj:erm}). $\mu(\epsilon)$ denotes an uncertainty probability measure for $\epsilon \in \mathbb{R}^d$ that can represent uniform balls such as $\|\epsilon\| \leq \rho$ (but other measures are possible as well). When $t \to \infty$ and $\mu(\epsilon)$ takes $\|\epsilon\|\leq \rho$, $L^t(\theta)$ reduces to the SAM objective $L^s(\theta)$. When $t = 0$, $L^t(\theta)$ reduces to the average loss over the perturbed neighborhood $\mu(\epsilon)$, i.e., $\mathbb{E}_{\mu(\epsilon)}[L(\theta+\epsilon)]$ where the expectation is taken with respect to the randomness of $\epsilon$ (formally proved in Appendix~\ref{app:proof:properties}). When we set $t=0$ and $\rho=0$, the TSAM loss is reduced to the classic average empirical risk $L(\theta)$. We may use $\mathbb{E}_{\mu(\epsilon)}$, $\mathbb{E}_{\epsilon}$, and $\mathbb{E}$ interchangeably when the meaning is clear from the context.

TSAM provides a smooth transition between min-max optimization (Eq.~(\ref{obj:sam})) and min-avg optimization $\min_{\theta} \mathbb{E}_{\mu(\epsilon)}[L(\theta+\epsilon)]$. {The min-avg optimization has appeared in prior works known as average-perturbed sharpness~\citep{wen2022does}, noise-perturbed loss~\citep{zhang2024noise}, or random smoothing~\citep{duchi2012randomized}.} The smoothness parameter of the TSAM  objective increases as the value of $t$ increases, which suggests that it is easier to optimize than SAM (Section~\ref{sec:properties}). As we formalize later, TSAM reweights gradients of neighboring solutions based on their loss values, which can be viewed as a soft version of SAM that assigns all the weights to one single worst minimum. 
In addition to the benefits in optimization, rigorously considering many, as opposed to one, neighbourhood parameters that incur large losses can result in improved generalization. We provide both theoretical characterization and empirical evidence showing that TSAM solutions are flatter than those of ERM and SAM.
One line of the mostly related works have explored tilted risks to reweight different data points~\citep{li2023tilted,robey2022probabilistically}. In this work, we use the TSAM framework to assign varying priority to local minima in the parameter space. 

To solve TSAM, we need to estimate the integral over $\mu(\epsilon)$ (Eq.~(\ref{obj:tsam})), or equivalently, to estimate the full gradient of the objective, which is a tilted aggregation of gradients evaluated at $L(\theta+\epsilon)$.  Both require sampling the perturbation $\epsilon$  with probability proportional to $e^{t L(\theta+\epsilon)}$ for the integration. Naively sampling $\epsilon$ at random to obtain $L^t(\theta)$ would be inefficient, as it is likely that $L(\theta+\epsilon)$ under the sampled $\epsilon$ is small and therefore we need many samples to converge to the true distribution. On the other hand, methods based on Hamiltonian Monte Carlo (HMC)~\citep{leimkuhler2004simulating} are guaranteed to arrive at the exact distribution. Inspired by the Euler's rules for HMC, we develop an algorithm to efficiently sample $\epsilon$'s and estimate the true gradient of $L^t(\theta)$. Our contributions are summarized as follows.

\paragraph{Contributions.} We propose TSAM, a new sharpness-aware optimization objective that reweights the parameters around local minima via exponential tilting. We rigorously study several properties of TSAM, showing that it always favors  flatter solutions as $t$ increases (Section~\ref{sec:properties}). To optimize TSAM, we adapt a specific HMC algorithm to efficiently sample the model perturbation $\epsilon$ (Section~\ref{sec:method}).  We empirically demonstrate that TSAM results in flatter solutions and superior generalization performance than SAM and its variants for deep neural networks including transformers on both image and text datasets (Section~\ref{sec:experiment}).

\section{Related Work}

\paragraph{Sharpness-Aware Minimization.} SAM regularizes overparameterized models by considering adversarial data points that have large training errors~\citep{foret2020sharpness,zheng2021regularizing}. The SAM variants, training dynamics, and applications in different models have been extensively studied in prior work~\citep{long2023sharpness, foret2020sharpness, bartlett2023dynamics,andriushchenko2022towards,chen2024does,kwon2021asam,chen2021vision,liu2022random,zhou2021sharpness,du2022sharpness,baek2024sam,zhao2022penalizing,mi2022make,zhuang2022surrogate,mueller2023normalization,pmlr-v235-xie24d,TahmasebiSBJJ24}. 
Notably, TSAM objective can be viewed as a special case of a general sharpness-aware approach~\citep{TahmasebiSBJJ24}, though~\citet{TahmasebiSBJJ24} do not specifically study TSAM.
Some work aim to improve efficiency of the SAM algorithm studying different relaxations~\citep{du2022sharpness,liu2022towards}. \citet{zhao2022penalizing} use a linear interpolation between normal gradients and SAM outer gradients evaluated at the max-loss parameter, which does not take into account the possibly many bad local minima for highly non-convex problems. \citet{zhou2021sharpness} perform sample-wise reweighting for SAM, as opposed to parameter-wise reweighting proposed herein. 
\citet{liu2022random} improve the inner max optimization by adding a random perturbation to the gradient ascent step to smoothen its trajectory. \citet{li2024enhancing} leverage a moving average of stochastic gradients in the ascent direction to reduce the gradient variance\footnote{Note that the notion of `variance’ in VASSO~\citep{li2024enhancing} refers to the variance of stochastic gradients compared with full gradients; whereas in TSAM, we examine the loss variance around the neighborhood regions, where the randomness in variance (Definition~\ref{def:weight_statistics}) comes from the perturbation $\epsilon$.}. Our goal is \textit{not} to better approximate the inner max or develop algorithms for solving the min-max SAM formulation, but rather, to solve a different TSAM objective that reweights many local minima to seek flat solutions. Nevertheless, we still compare with more advanced algorithms for the SAM objective and show the superiority of TSAM solutions (Section~\ref{sec:experiment}). As TSAM is a new objective, in principle, we can readily apply many existing optimization techniques (that can be potentially applied to SAM as well) such as variance reduction~\citep{johnson2013accelerating}, acceleration~\citep{nesterov1983method}, or adaptivity~\citep{streeter2010less,duchi2011adaptive,kingma2014adam} on top of the tilted stochastic gradients to gain further improvement.

There are also works exploring why SAM leads to better generalization or theoretically studying what SAM (and its implementation) is effectively minimizing~\citep{chen2024does,andriushchenko2022towards,long2023sharpness,wen2022does}. In this work, we prove that TSAM (and SAM) encourages flatter models for a class of problems including generalized linear models, where flatness (or sharpness) is characterized by the variance of the losses around the minima (Definition~\ref{def:sharpness}). Our proposed TSAM framework is particularly suitable for problems where flatness helps generalization. The various notions of sharpness, along with theoretical relations between sharpness and generalization still remain an open problem~\citep{andriushchenko2023modern,wen2024sharpness,ding2024flat,TahmasebiSBJJ24}, which is outside the scope of our paper.

\textbf{Tilting in Machine Learning.} Exponential tilting, used to shift parametric distributions, has appeared in previous literature in importance sampling, optimization, control, and information theory~\citep[e.g.,][]{siegmund1976importance,kort1972new,dembo2009large, term-generalization,whittle2002risk,whittle1981risk}. Recently, the idea of tilted risk minimization (which exponentially reweights different training samples) has been explored in machine learning applications such as enforcing fairness and robustness, image segmentation, and noisy label correction~\citep{li2023tilted,robey2022probabilistically,zhou2020robust,szabo2021tilted, term-generalization}. A closely-related LogSumExp operator is often used to as an smooth approximation to the max, which is always considered more computationally favorable~\citep{kort1972new,calafiore2014optimization,shen2010dual, li2023tilted}. One application of tilted risks applied to the adversarial training problem is to balance worst-case robustness (i.e., adversarial robustness) and average-case robustness in the data space~\citep{robey2022probabilistically}, among other approaches that can also achieve a transition between worst-case and average-case errors~\citep{rice2021robustness}. Our work is similar conceptually, but we consider reweighting adversarial model parameters, instead of adversarial data points. Compared with SAM (optimizing the largest loss), the TSAM framework offers additional flexibility of optimizing over quantiles of losses given the connections between exponential tilting and quantile approaches~\citep{rockafellar2000optimization,li2023tilted}. 

\section{Properties of TSAM} \label{sec:properties}

In this section, we discuss properties of the TSAM objective. We first state the convexity and smoothness of TSAM (Section~\ref{sec:properties:general}).  We then show that as $t$ increases, the gap between less-flat and more-flat solutions of the TSAM objective becomes larger. In other words, optimizing TSAM would give flatter solutions as $t$ increases (Section~\ref{sec:properties:glms}). Finally, we discuss the generalization behavior of TSAM and prove that there exists $t \in (0, \infty)$ that result in the tightest bound (Section~\ref{sec:properties:generalization}). All properties discussed in this section hold regardless of the distributions of $\epsilon$ (i.e., choice of $\mu(\epsilon)$), unless otherwise specified.

\subsection{Convexity and Smoothness} \label{sec:properties:general}

In this part, we connect the convexity and smoothness of TSAM with the convexity and smoothness of the ERM loss. We provide complete proofs in Appendix~\ref{app:proof:properties}.
We first define a useful quantity (tilted weights) that will be used throughout this section.
\begin{definition}[$t$-tilted weights] \label{def:weights}
    For a perturbed model parameter $\theta+\epsilon$, we define its corresponding $t$-tilted weight as 
        $w^t(\theta+\epsilon) := \frac{e^{t L(\theta+\epsilon)}}{\mathbb{E}[e^{tL(\theta+\epsilon)}]}$.
\end{definition}
The weight of parameter $\theta+\epsilon$ is exponentially proportional to the loss evaluated at this point. The expectation is with respect to the randomness of $\epsilon$ constrained by $\mu(\epsilon)$. When $t=0$, $0$-tilted weights are uniform. When $t \to \infty$, $w^t(\theta+\epsilon)$ focuses on the max loss among all possible $\{\theta+\epsilon\}$. Such weights have appeared in previous literature on importance sampling~\citep{siegmund1976importance}, but they are only applied to reweight sample-specific losses, as opposed to perturbation-specific parameters. Given tilted weights in Definition~\ref{def:weights}, we can present the TSAM gradients and Hessian as follows.
\begin{lemma}[Gradient and Hessian for TSAM]
Assume $L(\cdot)$ is continuously differentiable. 
The full gradient of TSAM (Objective~(\ref{obj:tsam})) is 
{\small
\begin{align}
    \nabla L^t(\theta) = \frac{\mathbb{E}[e^{t L(\theta+\epsilon)} \nabla 
 L(\theta+\epsilon)]}{\mathbb{E}[e^{t L(\theta+\epsilon)}]} = \mathbb{E}[w^t(\theta+\epsilon) \nabla L(\theta+\epsilon)]. \nonumber
\end{align}
}
The Hessian of TSAM $\nabla^2 L^t(\theta)$ is
\begin{align}
    & t \left(\mathbb{E}\left[w^t(\theta+\epsilon) \nabla L(\theta+\epsilon)^\top \nabla L(\theta+\epsilon)\right] \right. \nonumber \\ 
    & \left. - \mathbb{E}\left[w^t(\theta+\epsilon)  \nabla L(\theta+\epsilon)\right]^\top \mathbb{E}\left[w^t(\theta+\epsilon) \nabla L(\theta+\epsilon)\right] \right) \nonumber \\
     &+ \mathbb{E}\left[w^t(\theta+\epsilon)  \nabla^2 L(\theta+\epsilon)\right]. \label{eq:hessian}
\end{align}
\end{lemma}
The gradient of TSAM can be viewed as reweighting the gradients of $\nabla L(\theta+\epsilon)$ by the loss values $e^{tL(\theta+\epsilon)}$. Examining the Hessian, we note that the first term is $t$ multiplied by a positive semi-definite matrix, and the second term can be viewed as a reweighted Hessian of the original loss $L$ evaluted at $\theta+\epsilon$. 

It is not difficult to observe that if $L(\theta)$ is $p$-Lipschitz with respect to $\theta$, then $L^t(\theta)$ is $p$-Lipschitz with respect to $\theta$. If $L$ is $\mu$-strongly convex, then $L^t$ is also $\mu$-strongly convex (proof in Appendix~\ref{app:proof:properties}). Next, we show that the smoothness of TSAM scales linearly with $t$.
\begin{lemma}[Smoothness of TSAM] \label{lemma:smoothness}
    Let $L(\cdot)$ be $\beta$-smooth and $\beta$ is bounded. Then $L^t(\theta)$ is $\beta(t)$-smooth, where $\beta(t)$ satisfies $
        0 < \lim_{t \to \infty} \frac{\beta(t)}{t} < +\infty$.
\end{lemma}
That is, $\beta(t) = O(t)$. The proof is deferred to Appendix~\ref{app:proof:properties}.
In Lemma~\ref{lemma:smoothness}, we connect the smoothness of the tilted objective with the smoothness of the original ERM objective. We see that for any bounded $t$, the smoothness parameter is bounded.  As $t$ increases, TSAM becomes more difficult to optimize as the loss becomes more and more non-smooth.
When $t \to \infty$, $\beta(t) \to \infty$. If we have access to unbiased gradient estimates at each round, it directly follows that the convergence of SAM (TSAM with $t \to \infty$) objective is slower than that of tilted SAM following standard arguments~\citep{nesterov2013introductory}.  
To further visualize this, we create a one-dimensional toy problem in Appendix~\ref{app:toy}, where we obtain the globally optimal solutions for each objective. We show that both SAM and TSAM are able to arrive at flat solutions; but the SAM objective is non-smooth, hence more difficult to optimize.

\subsection{TSAM Prefers Flatter Models as $t$ Increases} \label{sec:properties:glms}

In this subsection, we focus on a specific class of models including generalized linear models (GLMs), where the loss function $l(x_i; \theta)$ carries the form of
{\setlength{\belowdisplayskip}{2pt} \setlength{\belowdisplayshortskip}{2pt}
\setlength{\abovedisplayskip}{2pt} \setlength{\abovedisplayshortskip}{2pt}
\begin{align}
&l(x_i;\theta) = A(\theta)-\theta^\top T(x_i),  \\
    &L(\theta) = A(\theta)-\theta^\top \left(\frac{1}{n}\sum_{i \in [n]} T(x_i)\right). \label{eq:linear_models}
\vspace{-0.1in}
\end{align} 
}

For GLMs, $A(\theta)$ is a convex function, and $\sum_{i \in [n]} T(x_i) T(x_i)^\top \succ 0$~\citep{wainwright2008graphical}. Our results in this section apply to loss functions defined in Eq.~(\ref{eq:linear_models}), which subsume linear models. Before introducing the main theorem, we define two important quantities that will be used throughout this section, and in the experiments.

\begin{definition}[$t$-weighted mean and variance] \label{def:weight_statistics}
We define $t$-weighted mean of a random variable $X$ as
$\frac{\mathbb{E}[e^{t X} X]}{\mathbb{E}[e^{t X}]}$.
Similarly, we define $t$-weighted variance of a random variable $X$ as 
$\frac{\mathbb{E}[e^{t X} X^2]}{\mathbb{E}[e^{t X}]} -\left(\frac{\mathbb{E}[e^{t X} X]}{\mathbb{E}[e^{t X}]}\right)^2$.
\end{definition}

When $t=0$, these definitions reduce to standard mean $\mathbb{E}[X]$ and variance $\mathbb{E}[X^2]-(\mathbb{E}[X])^2$. Similar tilted statistics definitions have also appeared in prior work~\citep{li2023tilted}. We leverage weighted variance to define sharpness below. 
\begin{definition}[$t$-sharpness] \label{def:sharpness}
    We say that a model parameter is $\theta_1$ is $t$-sharper than $\theta_2$ if the $t$-weighted variance of $L(\theta_1+\epsilon)$ (which is a random variable of loss distribution under model parameters perturbed by $\epsilon$) is larger than the   $t$-weighted variance of $L(\theta_2+\epsilon)$.
\end{definition}
Given the definition of sharpness above based on weighted variance, we are ready to prove that TSAM encourages  flatter local minima as the increase of $t$. Empirically, in Section~\ref{sec:experiment}, we also plot the $0$-sharpness of the solutions obtained from different objectives, and observe that TSAM achieves  smaller sharpness values (measured in terms of standard variance) than ERM and SAM. Proper definitions of sharpness is generally still an open problem, and other options are possible such as the trace of Hessian, gradient norms, and other functions over Hessian~\citep{wen2022does,TahmasebiSBJJ24}.

\begin{figure*}[t!]
    \centering
    \begin{subfigure}{0.22\textwidth}
        \centering
        \includegraphics[width=\textwidth]{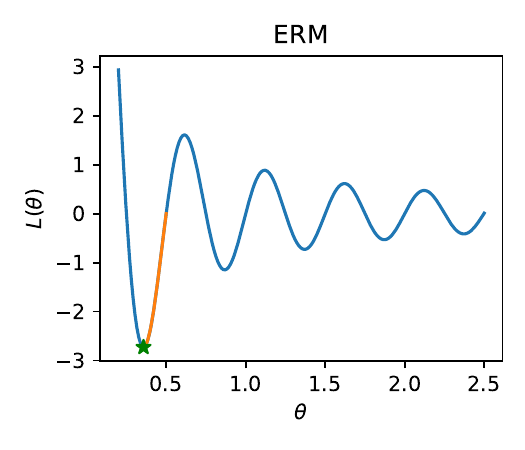}
    \end{subfigure}
    \hfill
    \begin{subfigure}{0.22\textwidth}
        \centering
        \includegraphics[width=\textwidth]{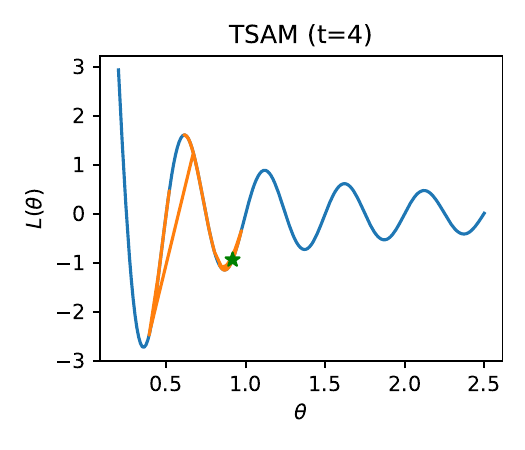}
    \end{subfigure}
    \hfill
    \vspace{-0.1in}
    \begin{subfigure}{0.22\textwidth}
        \centering
        \includegraphics[width=\textwidth]{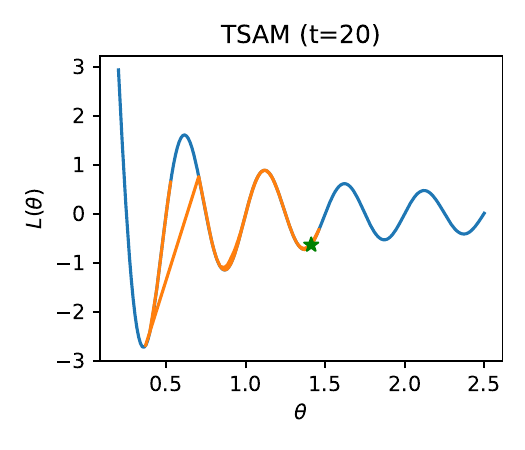}
    \end{subfigure}
    \hfill
    \begin{subfigure}{0.22\textwidth}
        \centering
        \includegraphics[width=\textwidth]{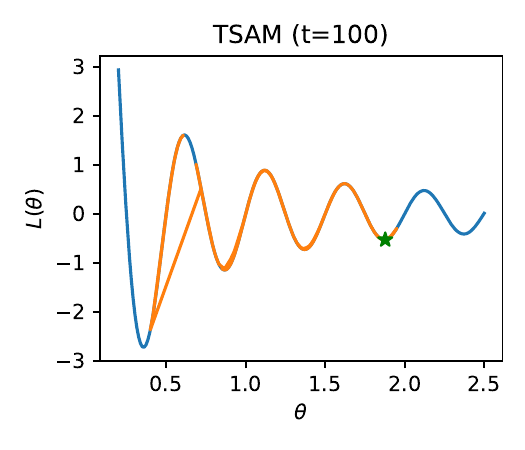}
    \end{subfigure}
    \caption{Optimization trajectories of different objectives (orange) with final solutions marked in star. The local minima gets flatter from left to right in each subfigure. We see that TSAM favors flat minima as $t$ increases. ERM solution by gradient descent converges to a sharp minimal. We solve TSAM on this one-dimensional problem by sampling thousands of $\epsilon$'s to estimate the gradient of $L^t(\theta)$  at each step, which is infeasible for real problems.}
\label{fig:toy}
\end{figure*}

\begin{theorem}[TSAM prefers  flatter models as $t$ increases] \label{thm:main}
Assume $L(\theta)$ is given by Eq.~(\ref{eq:linear_models}) and $L(\theta)$ is continuously differentiable.  For any $\theta_1, \theta_2 \in \mathbb{R}^d$, let $g^t(\theta_1, \theta_2) := L^t(\theta_1)-L^t(\theta_2)$. If $\theta_1$ is $t$-sharper than $\theta_2$, then $
       \frac{\partial g^t(\theta_1, \theta_2)}{\partial t} \geq 0.$
\end{theorem}

For some $\theta_1$ sharper than $\theta_2$, it is possible that $L(\theta_1)=L(\theta_2)$, which implies that ERM is not able to distinguish between the two solutions, while TSAM can. Furthermore,  Theorem~\ref{thm:main} indicates that as $t$ increases, the TSAM objective favors $\theta_2$ more aggressively, as the gap between $L^t(\theta_1)$ and $L^t(\theta_2)$ grows larger. We explore a one-dimensional toy problem with many local minima: $L(\theta)=2 \sin(4\pi  \theta)  / (2\theta) +  0.005 (\theta-1)^2$, and focus on the area $\theta \in (0.2, 2.5)$ to visualize this behavior. 
We take $\rho=0.2$ and a fixed learning rate $0.005$ for all objectives. Each run starts from the same initialization $\theta=0.5$. In Figure~\ref{fig:toy}, we see that as $t$ increases, TSAM leads to flatter solutions, despite having larger objective values measured by $L(\theta)$.  
As a side note, we prove that for any $\theta$, the objective value of $L^t(\theta)$ is monotonically increasing as $t$ increases (Appendix~\ref{app:proof:properties}). 
Next, we discuss a special case when $t$ is close to $0$, where we provide another perspective on the TSAM behavior.

\textbf{Discussions for $t \to 0$.} All the results above hold for the small-$t$ regime, where sharpness reduces to standard variance when $t \to 0$ (Definition~\ref{def:weight_statistics}). It still follows that $\left .\ \frac{\partial g^t (\theta_1, \theta_2)}{\partial t} \right |_{t \to 0}\geq 0$ if $\theta_1$ is sharper than $\theta_2$. Here, we provide another interpretation of TSAM when $t$ is close to zero. Similar statements have also appeared in prior works in a different context~\citep[e.g.,][]{liu2019deep,li2023tilted}. For a very small $t$, it holds that
$\frac{1}{t}\log \left(\mathbb{E}\left[e^{t L(\theta+\epsilon)}\right]\right) 	\approx \mathbb{E}[L(\theta+\epsilon)] + \frac{t}{2} \var\left(L(\theta+\epsilon)\right) + o(t^2).
$
We provide a proof in Appendix~\ref{app:proof:properties}.
Hence, optimizing TSAM is approximately equivalent to optimizing for the mean plus variance of the losses under the perturbed parameters. When $t=0$, it reduces to only optimizing for $\mathbb{E}[L(\theta+\epsilon)]$. 
In other words, TSAM with $t$ close to 0 is directly minimizing $0$-sharpness (standard variance).
For any $\theta_1$ and $\theta_2$ such that $\theta_1$ is sharper than $\theta_2$, we have
\begin{align*}
\vspace{-0.3in}
    g^t(\theta_1, \theta_2) &\approx \mathbb{E}[L(\theta_1+\epsilon)] + \frac{t}{2} \var \left(L(\theta_1+\epsilon)\right) \nonumber \\ &- \mathbb{E}[L(\theta_2+\epsilon)] - \frac{t}{2} \var \left(L(\theta_2+\epsilon)\right), \\ 
    \vspace{-0.3in}
    \frac{\partial g^t (\theta_1, \theta_2)}{\partial t} &\approx \frac{1}{2} \left(\var \left(L(\theta_1+\epsilon)\right) - \var \left(L(\theta_2+\epsilon)\right)\right) \geq 0.
\vspace{-0.1in}
\end{align*}
This is a special case of Theorem~\ref{thm:main} for $t \to 0$. It suggests that as we increase $t$ from 0 for a small amount, the standard variance of neighborhood loss would reduce. 
{Note that some recent works propose a noise-injected loss similar to TSAM with $t \to 0$~\citep{zhang2024noise}. The proposed algorithm therein explicitly minimizes the trace of Hessian, which aligns with our arguments that the TSAM objective can lead to flatter solutions. }

So far, we study properties of TSAM regarding convexity, smoothness of the objective, and sharpness of the resulting solutions. We note that these properties of the objective are independent of the actual optimization algorithms used to optimize TSAM. Though Theorem~\ref{thm:main} implies similar benefits of both TSAM and SAM relative to ERM (assuming we optimize TSAM and SAM perfectly), Lemma~\ref{lemma:smoothness} shows the superiority of the TSAM objective over SAM with unbounded smoothness parameters, as TSAM is easier to optimize. The performance of practical applications depends on both the properties of the objectives and the approximation algorithms used to solve them. 


\subsection{Generalization of TSAM} \label{sec:properties:generalization}
In this section, we give a uniform bound on the generalization error of our TSAM objective. By solving the tilted objective empirically, during test time, we are ultimately interested in evaluating the \textit{linear population risk} $\mathbb{E}_{Z}[l(\theta; Z)]$ where $Z$ denotes the underlying data distribution and $\{x_i\}_{i \in [n]} \sim Z$. We define  generalization error as the difference between population risk and our empirical objective value $\mathbb{E}_{Z} [l(\theta; Z)] - \frac{1}{t} \log \mathbb{E}_{\epsilon} [e^{t L(\theta+\epsilon)}]$, bounded as follows. 
\begin{theorem}[Generalization of TSAM] \label{thm:generalization}
Assume losses are bounded as $0 \leq L(\cdot) \leq M$. Suppose we have $n$ training data points. For any $\theta \in \Theta$ and $t \geq 0$, with probability $1-\delta$, the difference between population risk and empirical TSAM risk $\text{gen}_{t} :=  \mathbb{E}_{Z} [l(\theta; Z)] - \frac{1}{t} \log \mathbb{E}_{\epsilon} [e^{t L(\theta+\epsilon)}]  $ satisfies
    \begin{align}
         \text{gen}_{t} \leq M \sqrt{\frac{\log(2/\delta)}{2n}} -\frac{\var_{\epsilon} (e^{tL(\theta+\epsilon)})}{2t e^{2tM}} + c, \label{eq: gen1}
    \end{align}
    where $c = L(\theta) - \mathbb{E}_{\epsilon}[L(\theta+\epsilon)]$ is a constant unrelated to $t$.
\end{theorem}

We defer the proof to Appendix~\ref{app:proof:properties}, where we build upon existing generalization results of a related objective~\citep{term-generalization}. From Theorem~\ref{thm:generalization}, we see that when the sample space of $\epsilon$ is empty, our result reduces to $\mathbb{E}_{Z} [l(\theta, Z)] \leq L(\theta) + M \sqrt{\frac{\log(2/\delta)}{2n}}$, scaling at a rate of $\frac{1}{\sqrt{n}}$ consistent with standard uniform bound on the average risk~\citep{shalev2014understanding}. 
When $t \to \infty$ and we define $\mu(\epsilon)$ to be $\|\epsilon\|\leq \rho$ over some distribution, {the result gives an upper bound on the generalization of SAM: $\mathbb{E}_Z[l(\theta, Z)]-\max_{\|\epsilon\|\leq \rho} L(\theta+\epsilon) \leq M \sqrt{\frac{\log(2/\delta)}{2n}}+c$.} 

Additionally, denote $\theta^{\text{TSAM}}$ and $\theta^{\text{ERM}}$ as optimal solutions for TSAM (Eq.~(\ref{obj:tsam})) and ERM (Eq.~(\ref{obj:erm})), respectively. For modest values of $t$, due to the negativity of $-\frac{\var_{\epsilon} (e^{tL(\theta+\epsilon)})}{2t e^{2tM}}$, the upper bound of the linear population risk $\mathbb{E}_Z[l(\theta^{\text{TSAM}}, Z)]$   can be smaller than that of the linear risk $\mathbb{E}_Z[l(\theta^{\text{ERM}}, Z)]$, as long as $\frac{1}{t} \log \mathbb{E}_{\epsilon} [e^{t L(\theta^{\text{TSAM}}+\epsilon)}] -\frac{\var_{\epsilon} (e^{tL(\theta^{\text{TSAM}}+\epsilon)})}{2t e^{2tM}} \leq L(\theta^{\text{ERM}})$. This implies that by solving TSAM, we can obtain a solution that results in a smaller {upper bound of the} linear population error than that of ERM. 

\section{Algorithms} \label{sec:method}

In this section, we describe the algorithms we use to solve TSAM. The main challenge in solving TSAM is to sample $\epsilon$ to get a good estimator of $L^t(\theta)$, or equivalently, $\nabla L^t(\theta)$. We first describe a general approach where we use estimated tilted gradients (given sampled $\epsilon$'s) to update the model (Section~\ref{sec:method:hmc}). 
Then, we discuss how to sample $\epsilon$'s via a specific Hamiltonian Monte Carlo algorithm and present our method and implementation (Section~\ref{sec:method:ours}).


\subsection{General Algorithm} \label{sec:method:hmc}

To solve TSAM, the primary challenge is to estimate the integral $\frac{1}{t} \log \left(\int e^{tL(\theta+\epsilon)} d \mu(\epsilon) \right)$, or its full gradient $\frac{\mathbb{E}[e^{tL(\theta+\epsilon)} \nabla L(\theta+\epsilon)]}{\mathbb{E}[e^{tL(\theta+\epsilon)}]}$, assuming  gradient-based methods and the differentiable loss $L$. A naive way is to first sample $\epsilon$ from $\mu(\epsilon)$ following the pre-defined distribution (e.g., Gaussian or uniform) over $\mu(\epsilon)$, and then perform tilted aggregation with weights proportional to $e^{t L(\theta+\epsilon)}$. However, this approach may be extremely inefficient, as there could be an infinite set of perturbed model parameters with relatively small losses, which are not informative. In Figure~\ref{fig:compare_random_sampling} in the appendix, we empirically show that even when we sample a much larger number of $\epsilon$'s, the resulting accuracy is still worse than our proposed method. Instead, we propose to 
sample $s$ number of $\epsilon$'s from distribution $e^{\delta L(\theta+\epsilon)}$ (denoted as $\{\epsilon_j\}_{j \in [s]}$), where $0 \leq \delta \leq t$.  We then use these $\{\epsilon_j\}_{j \in [s]}$ to obtain an empirical gradient estimation with weights proportional to $\{e^{(t-\delta) L(\theta+\epsilon_j)}\}_{j \in [s]}$, as the full gradient is a tilted average of the original gradient on $L(\cdot)$. 
To improve sample efficiency, we use gradient-based methods such as Hamiltonian Monte Carlo (HMC) that simulates Hamiltonian dynamics~\citep{leimkuhler2004simulating}. The structure of our proposed method is in Algorithm~\ref{alg:sampling}. 
Note that in principle, after estimating the tilted stochastic gradients, we can further apply existing optimization techniques such as variance reduction~\citep{johnson2013accelerating}, acceleration~\citep{nesterov269method}, or adaptivity~\citep{streeter2010less,duchi2011adaptive}  to gain further improvement, which we leave for future work.

\setlength{\textfloatsep}{2pt}
\begin{algorithm}[h!]
\caption{Tilted SAM Solver}
\begin{algorithmic}[1]
\REQUIRE {$t$, $\theta^0$, learning rate $\eta$, total iterations $T$, total number of samples $s$, $0 \leq \delta \leq t$}
\label{alg:sampling}
    \FOR{$i=0, \cdots, T-1$}
            \STATE Sample $s$ random perturbations $\{\epsilon_j\}_{j \in [s]}$ from distribution $e^{\delta L(\theta^i+\epsilon)}$ under the constraint characterized by $\mu(\epsilon)$ via some HMC algorithm (Algorithm~\ref{alg:main})\; 
            \STATE Update $\theta^i$ with the estimated gradient evaluated on the mini-batch:
            \setlength{\abovedisplayskip}{5pt}
            \setlength{\belowdisplayskip}{2pt}
            \[
                \theta^{i+1} \leftarrow \theta^i - \eta \frac{\sum_{j \in [s]} e^{(t-\delta) L\left(\theta^i + \epsilon_j\right)} \nabla L(\theta^i + \epsilon_j)}{\sum_{j \in [s]} 
 e^{(t-\delta) L\left(\theta^i+\epsilon_j\right)}}
            \]
    \ENDFOR
\STATE \textbf{Return} $\theta^T$
\end{algorithmic}
\end{algorithm}

\subsection{Sampling $\epsilon$} \label{sec:method:ours}

There could be potentially different algorithms for sampling $\epsilon$ where $p(\epsilon)  ~\propto  ~e^{\delta L(\theta+\epsilon)}$.
Here we propose an approximate and cheap sampler based on discretization of Hamiltonian dynamics.
Our method is inspired by one of the best-known way to approximate the solution to a system of differential equations, i.e., Euler’s method or its modification~\citep{neal2011mcmc}.
A more accurate solver like the leap-frog method might be more popular for HMC, but these come at an increased expense~\citep{neal2011mcmc}.
As our goal to minimize computational cost, we stick with the cheaper Euler's approach as follows.
We first initialize $\epsilon_0$ from an $L_2$ ball that satisfies $\|\epsilon\| \leq \rho$, and initialize the momentum $p_0 \in \mathbb{R}^d$ from some Gaussian distribution, i.e., $p_0 \sim \mathcal{N}(0, \sigma^2 \mathbf{I})$. Note that the negative log probability density of the energy function $U(\epsilon)$ is $-\log(e^{\delta L(\theta+\epsilon)}) = -\delta L(\theta+\epsilon)$.
At each sampling step, we run the following steps for $N$ iterations with a small step-size $\beta$ to obtain a candidate $\epsilon$:
\begin{align}
    p \leftarrow p + \beta \delta \nabla_{\epsilon} L(\theta+\epsilon),  \quad
    \epsilon \leftarrow \epsilon + \beta p / \sigma^2. \label{eq:hmc_2}
\end{align}

After obtaining a candidate $\epsilon$, we accept $\epsilon$ with probability $\min\{1, e^{ \delta 
 L(\theta+\epsilon) - \frac{\|p\|^2}{2\sigma^2}}  / e^{\delta L(\theta+\epsilon_0)-\frac{\|p_0\|^2}{2\sigma^2}}\}$. If the candidate $\epsilon$ is not accepted, we set $(p,\epsilon)$ to the initial point before the $N$ iterations.
Repeating the above for enough times would give us a sample $\epsilon$ from the exact distribution.

\begin{algorithm}[h!]
\caption{Tilted SAM Solver}
\begin{algorithmic}[1]
\REQUIRE{$\theta^0$, total samples $s$, uncertainty ball radius $\rho$}
\caption{Sampling from $e^{\delta L(\theta^i+\epsilon)}$ where $\|\epsilon\| \leq \rho$}
\label{alg:main}
        \FOR{$j=0, \cdots, s$}
            \STATE Perturb $\theta^i$ with a random $\delta_j$ sampled from Gaussian or uniform distribution: $\theta_j^i \leftarrow \theta^i + \delta_j$ \; 
            \STATE Run normalized SGD on the mini-batch data at $\theta_j^i$:  
            \setlength{\abovedisplayskip}{5pt}
            \setlength{\belowdisplayskip}{2pt}
            \[ \hat{\theta}_j^i \leftarrow \theta_j^i +  \rho \frac{\nabla L(\theta^i_j)}{\|\nabla L(\theta^i_j)\|}; ~~
            \epsilon_j \leftarrow \hat{\theta}_j^i -\theta^i
            \]
        \ENDFOR
    \STATE \textbf{Return} {$\{\epsilon_j\}_{j \in [s]}$} 
\end{algorithmic}
\end{algorithm}

Generating one $\epsilon$ via HMC requires at least $2N$ gradient evaluations, which is infeasible for large-scale problems. Hence, we set $N=1$ {in all the main experiments}, and meanwhile accept the generated $\epsilon$ with probability 1. {For completeness, we evaluate the effects of increasing $N$ in HMC in Appendix~\ref{app:experiments}. We observe that using $N>1$ does not significantly improve the performance. Running equations in Eq.~(\ref{eq:hmc_2}) for one step,} if $p$ is initialized as $p = \mathbf{0}$, we have $\epsilon \leftarrow \epsilon + \beta' \nabla L(\theta+\epsilon)$, where $\beta'$ is a constant. We adopt this updating rule in our problem, and run the aforementioned procedure in parallel for $s$ times to get $s$ samples. Our method is presented in Algorithm~\ref{alg:main}. 
Though Algorithm~\ref{alg:main} does not guarantee the $\epsilon_j$'s result in a consistent estimator of the TSAM integral, we empirically showcase its effectiveness on non-convex models including transformers in the next section. 



\section{Experiments} \label{sec:experiment}

In this section, we first describe our setup. Then we present our main results, comparing TSAM with the baselines of ERM (Eq.~(\ref{obj:erm})), SAM (Eq.~(\ref{obj:sam})), and SAM variants on both image and text data (Section~\ref{sec:experiment:main}). We explain TSAM's superior performance by empirically examining the flatness of local minima in Section~\ref{sec:experiments:sharpness}. In Section~\ref{sec:experiments:hyperparameter}, we discuss the effects of hyperparameters. 
Our code is publicly available at \href{https://github.com/litian96/TSAM}{github.com/litian96/TSAM}.

\begin{table*}[!t]
\caption{TSAM achieves higher test performance relative to ERM and different variants of SAM  across  image datasets and the GLUE benchmark with both CNNs and transformers. TSAM (or SAM) is particularly suitable for applications with distribution shifts (DTD and noisy CIFAR100 datasets), which is also consistent with observations in prior works~\citep{foret2020sharpness,baek2024sam}. TSAM also results in lower test loss, discussed in detail in the next section. 
}
\vspace{1em}
\centering
\label{table:main}
\setlength{\tabcolsep}{6pt}
\scalebox{0.95}{
\begin{tabular}{lcc|cc|cc|cc}
	   \toprule[\heavyrulewidth]
        \multirow{2}{*}{\textbf{Methods}} & \multicolumn{2}{c}{\textbf{CIFAR100}}  &  \multicolumn{2}{c}{\textbf{DTD}}  & \multicolumn{2}{c}{\textbf{Noisy CIFAR100}} &  \multicolumn{2}{c}{\textbf{TinyImagenet}}   \\
        & ResNet18 & WideResNet & ViT & WideResNet & ResNet18 & WideResNet & ResNet18 & ResNet34 \\
        \midrule
     ERM & 71.39 {\small (.2)} & 73.22 {\small (.2)} & 66.38 {\small (.2)} & 16.97 {\small (.3)} & 61.01 {\small (.2)} & 57.03 {\small (.3)} & 71.10  {\small (.2)} & 74.90 {\small (.1)} \\
     SAM & 76.52 {\small (.1)} & 78.44 {\small (.1)} & 67.87 {\small (.2)} & 17.45 {\small (.2)} & 69.00 {\small (.1)} & 68.02 {\small (.2)}  & 72.43 {\small (.1)} & 76.75 {\small (.2)} \\
     ESAM1 & 77.40 {\small (.1)} & 80.22 {\small (.1)} & 68.18 {\small (.2)} & 17.67 {\small (.2)} & 69.20 {\small (.1)} & 69.79 {\small (.1)}  & 73.24 {\small (.1)} & 77.41 {\small (.1)} \\
     ESAM2 & 77.52 {\small (.1)}  & 79.03 {\small (.2)} & 68.35 {\small (.1)} & 17.71 {\small (.1)}  & 67.27 {\small (.1)} & 66.83 {\small (.1)} & 73.26 {\small (.1)} & 77.40 {\small (.1)} \\
     PGN & 77.45 {\small (.1)} &  78.58 {\small (.2)} & 67.76 {\small (.1)} & 18.23 {\small (.2)}   & 65.68 {\small (.1)}& 64.02 {\small (.2)} & 73.18 {\small (.1)} & \textbf{77.80} {\small (.1)} \\
     RSAM & 77.35 {\small (.1)}  & 79.02 {\small (.2)} & 68.35 {\small (.2)} & 17.66 {\small (.3)}  & 69.31 {\small (.1)} & 65.93 {\small (.2)} & \textbf{73.57} {\small (.1)} & \textbf{77.72} {\small (.1)} \\
     TSAM & \textbf{77.78} {\small (.1)}  & \textbf{80.85} {\small (.2)} & \textbf{68.82} {\small (.1)} & \textbf{18.63} {\small (.2)}  & \textbf{69.98} {\small (.1)} & \textbf{70.26} {\small (.1)} & \textbf{73.55} {\small (.1)} & \textbf{77.79} {\small (.1)} \\
\bottomrule[\heavyrulewidth]
\end{tabular}}

\centering
\scalebox{0.97}{
\begin{tabular}{lcccccccccc}
	   \toprule[\heavyrulewidth]
        \multicolumn{1}{l}{\textbf{Objectives}} & CoLA & WNLI & SST-2 & MNLI
 & QNLI & RTE & MRPC & QQP & STSB & \textbf{AVG} \\
        \midrule
    ERM &  52/80.34 &  54.93 & 90.48 & 79.6 & \textbf{87.72} & \textbf{60.65} & 83.82 & 86.32 & 86.6/86.3 & 77.15 \\
    SAM &  52/80.48 & \textbf{56.34} & \textbf{91.74} & \textbf{81.1} & 86.42 & 58.84 & {85.29} & 87.71 & \textbf{87.0}/86.5 & 77.56 \\
    {ESAM1} & {52/80.44} &	{\textbf{56.34}}	& {91.63}	&{\textbf{81.1}}&	{86.18}&	{59.02}&	{85.31}&	{87.69}&	{\textbf{87.1}/86.7}	& {77.59} \\
    {ESAM2}  & {52/80.53}	& {\textbf{56.34}} &	{91.63}	& {\textbf{81.2}}	& {86.29}	& {59.25}	& {\textbf{85.80}}	& {87.47}	& {86.8/86.5}	& {77.62} \\
   TSAM &  52/\textbf{80.81} &  \textbf{56.34} & \textbf{91.86} & \textbf{81.1} & \textbf{87.81} & \textbf{60.65} & 85.05 & \textbf{88.77} & \textbf{87.1}/86.6 & \textbf{78.01} \\
\bottomrule[\heavyrulewidth]
\end{tabular}}
\end{table*}


\textbf{Tasks and Datasets.} 
We study image classification tasks with CNNs and transformers, and language modeling tasks with BERT. 
First, we explore training ResNet18~\citep{he2016deep} {and WideResNet16-8~\citep{zagoruyko2016wide}} on CIFAR100~\citep{krizhevsky2009learning}. Previous works show that SAM is robust to label noise~\citep{foret2020sharpness,baek2024sam}; and we investigate this setting by training the same models on on CIFAR100 with label noise generated by substituting 20\% of the true labels uniformly at random to other labels.
Since vision transformers (ViTs)~\citep{dosovitskiy2020image} have been shown to result in sharper local minima than CNNs~\citep{chen2021vision}, we study the performance of finetuning ViTs (pretrained on ImageNet~\citep{deng2009imagenet}) on an out-of-distribution Describable Texture Dataset (DTD)~\citep{cimpoi14describing}, where the task is 47-class classification.  {We also use WideResNet16-8 to train DTD from scratch.} Additionally, we evaluate a 200-class classification task for TinyImagenet~\cite{le2015tiny} with ResNet18 and ResNet34~\citep{he2016deep} models.
Lastly, for text data, we study finetuning a pretrained DistilBERT~\citep{sanh2019distilbert} model on the GLUE benchmark including both classification and regression problems. 

\textbf{Hyperparameter Tuning.} We take $\mu(\epsilon)$ to be $\|\epsilon\| \leq \rho$ for all TSAM experiments, and tune the $\rho$ parameters separately from $\{0.05, 0.1, 0.2\}$ for relevant methods.
For TSAM, we tune $t$ from $\{0, 1, 5, 20, 100\}$ and select the best one based on the validation set. 
We also report the performance for all $t$'s in the next sections.
We use $s$=3 or $s$=5 sampled $\epsilon$'s for all datasets and find that it works well. 
For some SAM variants that introduce additional hyperparameters, we tune those via grid search as well. 
The batch size is 64 for all the datasets and methods and a constant learning rate is tuned from $\{0.0003, 0.001, 0.003, 0.01, 0.03, 0.1\}$ for each algorithm.  
Despite the existence of adaptive methods for SGD and SAM~\citep{kingma2014adam,kwon2021asam}, we do not use adaptivity for any algorithm for a fair comparison. 
See Appendix~\ref{app:experiments} for details on hyperparameter tuning.

\subsection{TSAM Leads to Better Test Performance} \label{sec:experiment:main}
We compare the performance of various objectives and algorithms in Table~\ref{table:main}.  ERM denotes minimizing the empirical average loss with mini-batch SGD. SAM is the vanilla SAM implementation with one step of gradient ascent and one step of gradient descent at each iteration~\citep{foret2020sharpness}. 
Note that TSAM requires more gradient evaluations per iteration. Hence, we include two additional baselines of SAM \textit{under the same computational budget as TSAM} runs. (1) We simply run the vanilla SAM algorithm for more iterations until it reaches the same runtime as TSAM. (2) We try another SAM approximation by exploring different step sizes along the gradient ascent directions and pick the one incurring the biggest loss. Then we evaluate the gradient under that step size to be applied to the original model parameters. 
We call these expensive SAM baselines ESAM1, and ESAM2, respectively. We also evaluate two more advanced sharpness-aware optimization methods: PGN that combines normal gradients and SAM gradients~\citep{zhao2022penalizing}, and Ramdom SAM (RSAM) which adds random perturbations before finding the adversarial directions~\citep{liu2022random}. \textit{We let PGN and RSAM run the same amount of time as TSAM} on the same computing platform. On all the datasets, we tuned $t$ values via grid search from $\{0, 1, 5, 20, 100\}$. 

Our results are shown in Table~\ref{table:main}.
The performance for all $t$'s on three image datasets {and different model architectures} are reported in Section~\ref{sec:experiments:hyperparameter}.
For the GLUE benchmark, we report the standard metrics for each dataset in GLUE.
TSAM consistently achieves higher test performance than ERM and variants of SAM. 
We provide corresponding convergence plots of ERM, vanilla SAM, and TSAM in Appendix~\ref{app:experiments}.


\subsection{Flatness of TSAM Solutions}
\label{sec:experiments:sharpness}

\begin{figure}[h!]
    \raggedright
    \begin{subfigure}{0.23\textwidth}
        \centering
        \includegraphics[width=\textwidth]{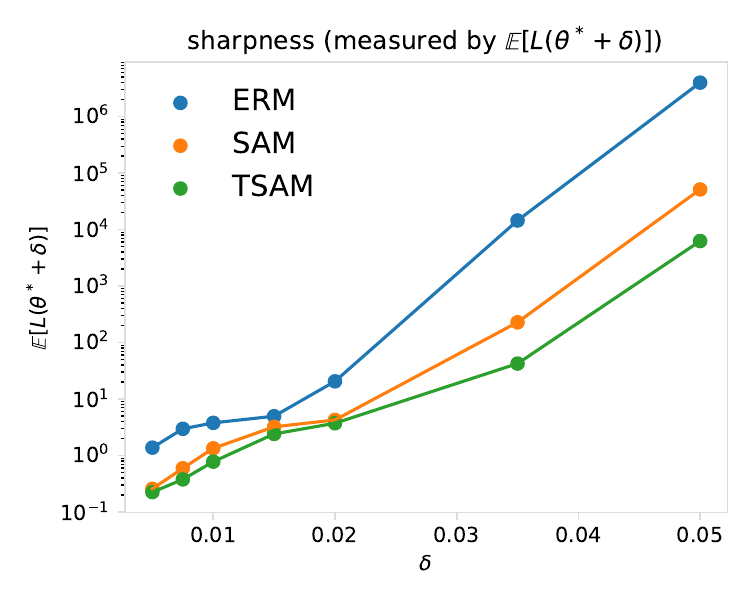}
    \end{subfigure}
    \hfill
    \begin{subfigure}{0.23\textwidth}
        \centering
        \includegraphics[width=\textwidth]{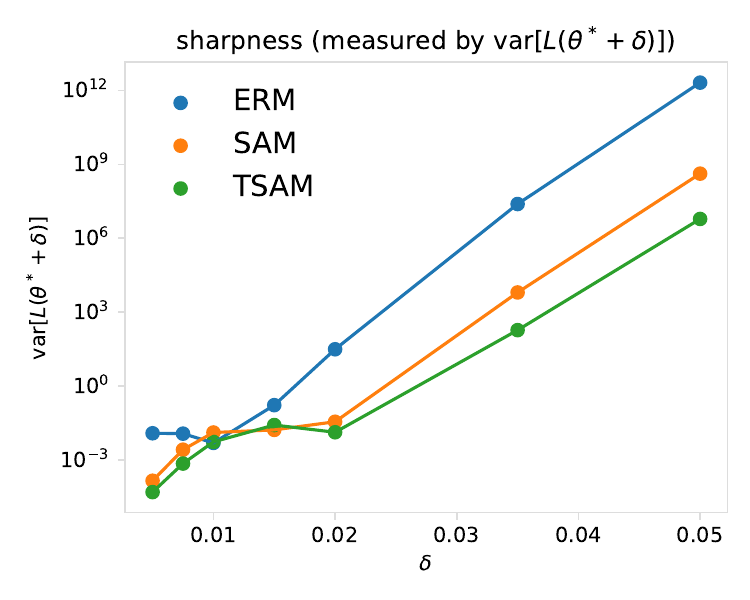}
    \end{subfigure}
    \caption{Sharpness of the solutions found by ERM, SAM, and TSAM on CIFAR100 with ResNet18. We empirically measure sharpness by both $\mathbb{E}[L(\theta^*+\epsilon)]$ and $\var[L(\theta^*+\epsilon)]$ where $\epsilon \sim \mathcal{N}(0, \delta^2)$. $\theta^*$ denotes different optimal model parameters obtained from the three objectives.  These metrics (especially variance) are also consistent with Definition~\ref{def:sharpness} with $t=0$. We see that TSAM solutions have a  flatter neighborhood compared with the other two.}
    \vspace{0.1in}
\label{fig:sharpness}
\end{figure}

In this part, we take a more detailed look into the properties of TSAM solutions compared with the ones of ERM and SAM on the CIFAR100 dataset trained by ResNet18 from scratch. In Figure~\ref{fig:sharpness}, we plot the loss mean and variance over the neighborhood areas around local minima obtained by different objectives, i.e., $\mathbb{E}[L(\theta^*+\epsilon)]$ and $\var[L(\theta^*+\epsilon)]$, where $\epsilon \sim \mathcal{N}(0, \delta^2)$, and $\theta^*$ denotes the different solutions of any objective (with a slight abuse of notation). These measurements have appeared in prior works {named average-loss sharpness}~\citep{wen2024sharpness, chen2021vision}, and are consistent with our sharpness definition (Definition~\ref{def:sharpness}) mentioned before. In Figure~\ref{fig:sharpness}, for all $\delta$ values, we see that TSAM consistently result in  flatter local minima than ERM and SAM measured by both the mean and variance of losses around the minima. {In addition, we evaluate sharpness following other common notions by investigating the top-5 eigen values of Hessian~\citep[e.g.,][]{foret2020sharpness}. Under the same model setup, the top-5 eigenvalues are \{342.11, 304.72, 260.71, 252.92, 210.88\} for ERM, \{232.60, 198.35, 182.61, 153.74, 145.76\} for SAM, and \{140.91, 113.38, 105.90,  92.94,  89.55\} for TSAM ($t$=20). We see that TSAM achieves the smallest max eigenvalues among the three.}

We further report the training and test performance of best-tuned ERM, SAM, and TSAM  in Table~\ref{table:train_test} in the appendix. We show that ERM solutions have lower training losses but higher test losses than SAM and TSAM when evaluated on the average test performance (i.e., the `ERM' column in the right table). This is due to the fact that ERM does not generalize as well as SAM or TSAM, and there exist bad sharp local minima around ERM solutions. On the other hand, while TSAM's average training loss is the highest (which is expected because it does not directly optimize over ERM), the test losses of TSAM for both the average-case performance and worst-case performance are lower than the other two baselines. {While we show better generalization of TSAM empirically, rigorous understandings between generalization and flatness remains an open area of research.}

\subsection{Sensitivity to Hyperparameters} \label{sec:experiments:hyperparameter}



\paragraph{Effects of the Tilting Hyperparameter $t$.} One critical hyperparameter in TSAM is $t$. When $t=0$, TSAM objective reduces to {the average-case perturbed objective}. When $t \to \infty$, the TSAM objective (Eq.~(\ref{obj:tsam})) recovers SAM (Eq.~\ref{obj:sam}). But the TSAM algorithm (Algorithm~\ref{alg:main}) do not exactly recover SAM's alternating updating approximation when $t \to \infty$. See Section~\ref{sec:method} for a detailed discussion.  Here, we report the test accuracies as the training proceeds under multiple values of $t$'s for all the three tasks. Results are plotted in Figure~\ref{fig:multiple_t} in the appendix. We see that there are a range of $t$'s that result in faster convergence or higher accuracies than SAM.  There also exists an optimal $t$ that leads to the best test performance. This is consistent with our previous generalization bound (Section~\ref{sec:properties:generalization}). Though Theorem~\ref{thm:main} captures the benefits of both SAM and TSAM, we note that the final empirical performance does not only depend on the properties of the objectives. But rather, it also relies on the choice of approximation algorithms. Results in Theorem~\ref{thm:main} assume that the objectives are optimized perfectly, which is infeasible in high-dimensional settings. 


Additionally, we empirically study the effects of scheduling $t$ during optimization. Increasing $t$ from 0 to a fixed value effectively switches from weighting local minima uniformly to rewighting them based on the loss values, and vice versa. We experiment with two options: linearly decreasing $t$ and linearly increasing $t$ on the noisy CIFAR100 dataset trained by ResNet18. The convergence curves are shown in Figure~\ref{fig:schedult_t}. We see that using a fixed $t$ throughout training does not have significant difference from scheduling $t$. Hence, we stick to fixed $t$'s for our TSAM experiments.


\begin{table}[h!]
    \centering
    \setlength{\tabcolsep}{4pt}
    \scalebox{0.98}{
    \begin{tabular}{lccc}\\
    \toprule  
    & CIFAR100 & DTD & noisy CIFAR100 \\
    \midrule
    SAM & 76.52 & 67.87 & 69.00 \\  
    TSAM, $s$=3 & 77.40 & \textbf{68.82} & 69.55 \\  
    TSAM, $s$=5 & \textbf{77.78} & \textbf{68.70} & \textbf{69.98} \\  \bottomrule
    \end{tabular}}
    \caption{Test accuracies of SAM and TSAM with different number of sampled $\epsilon$'s (denoted as $s$; see Algorithm~\ref{alg:main}). Empirically, we do not need many samples from the tilted distribution.}
    \vspace{0.2in}
    \label{table:s}
\end{table}


\paragraph{Effects of the Number of $\epsilon$'s.} 
One may wonder if we need to sample a large number of perturbations for the algorithm to be effective.
In Table~\ref{table:s}, we show that we usually only need $s=3$ or 5 number of $\epsilon$'s to achieve significant improvements relative to SAM. The models in three columns correspond to ResNet18, ViT, and ResNet18, respectively.

\vspace{0.4in}
\section{Conclusion} \label{sec:conclusion}

In this work, we have proposed a tilted sharpness-aware minimization (TSAM) objective, which leverages exponential tilting (parameterized by $t$) to reweight potentially many local minima in the neighborhoods. TSAM is a more smooth problem relative to SAM with a bounded $t$, and it explicitly encourages flatter solutions as $t$ increases. We have proposed a practical algorithm motivated by HMC to sample from the tilted distribution $e^{tL(\theta+\epsilon)}$. Through experiments on different models and datasets including label-noise settings, we have demonstrated that TSAM consistently outperforms SAM and its variants on both image and text datasets. 

\section*{Acknowledgements}

We thank Ahmad Beirami, Behrooz Tahmasebi, and Manzil Zaheer for helpful discussions. We thank the anonymous reviewers for their feedback.

\section*{Impact Statement}

In this work, we propose tilted sharpness-aware minimization (TSAM) that leverages exponential tilting to reweight sharp local minima and achieve better generalization. It could potentially improve the test performance of a wide range of models including those designed to address societal problems such as fairness, robustness, and privacy.  As the algorithmic framework is general, we do not anticipate our objective or algorithm to directly cause any negative impacts. However, if TSAM is applied in an undesirable manner, e.g., to some malicious attacking approaches, it may strengthen the capabilities of those models.

\nocite{langley00}

\bibliography{ref}

\begin{thebibliography}{61}
\providecommand{\natexlab}[1]{#1}
\providecommand{\url}[1]{\texttt{#1}}
\expandafter\ifx\csname urlstyle\endcsname\relax
  \providecommand{\doi}[1]{doi: #1}\else
  \providecommand{\doi}{doi: \begingroup \urlstyle{rm}\Url}\fi

\bibitem[Aminian et~al.(2024)Aminian, Asadi, Li, Beirami, Reinert, and Cohen]{term-generalization}
Aminian, G., Asadi, A.~R., Li, T., Beirami, A., Reinert, G., and Cohen, S.~N.
\newblock Generalization error of the tilted empirical risk.
\newblock \emph{arXiv preprint arXiv:2409.19431}, 2024.

\bibitem[Andriushchenko \& Flammarion(2022)Andriushchenko and Flammarion]{andriushchenko2022towards}
Andriushchenko, M. and Flammarion, N.
\newblock Towards understanding sharpness-aware minimization.
\newblock In \emph{International Conference on Machine Learning}, pp.\  639--668. PMLR, 2022.

\bibitem[Andriushchenko et~al.(2023)Andriushchenko, Croce, M{\"u}ller, Hein, and Flammarion]{andriushchenko2023modern}
Andriushchenko, M., Croce, F., M{\"u}ller, M., Hein, M., and Flammarion, N.
\newblock A modern look at the relationship between sharpness and generalization.
\newblock \emph{arXiv preprint arXiv:2302.07011}, 2023.

\bibitem[Baek et~al.(2024)Baek, Kolter, and Raghunathan]{baek2024sam}
Baek, C., Kolter, Z., and Raghunathan, A.
\newblock Why is sam robust to label noise?
\newblock \emph{arXiv preprint arXiv:2405.03676}, 2024.

\bibitem[Bartlett et~al.(2023)Bartlett, Long, and Bousquet]{bartlett2023dynamics}
Bartlett, P.~L., Long, P.~M., and Bousquet, O.
\newblock The dynamics of sharpness-aware minimization: Bouncing across ravines and drifting towards wide minima.
\newblock \emph{Journal of Machine Learning Research}, 24\penalty0 (316):\penalty0 1--36, 2023.

\bibitem[Boucheron et~al.(2013)Boucheron, Lugosi, and Massart]{10.1093/acprof:oso/9780199535255.001.0001}
Boucheron, S., Lugosi, G., and Massart, P.
\newblock \emph{{Concentration Inequalities: A Nonasymptotic Theory of Independence}}.
\newblock Oxford University Press, 2013.

\bibitem[Calafiore \& El~Ghaoui(2014)Calafiore and El~Ghaoui]{calafiore2014optimization}
Calafiore, G.~C. and El~Ghaoui, L.
\newblock \emph{Optimization Models}.
\newblock Cambridge University Press, 2014.

\bibitem[Chen et~al.(2021)Chen, Hsieh, and Gong]{chen2021vision}
Chen, X., Hsieh, C.-J., and Gong, B.
\newblock When vision transformers outperform resnets without pre-training or strong data augmentations.
\newblock \emph{arXiv preprint arXiv:2106.01548}, 2021.

\bibitem[Chen et~al.(2024)Chen, Zhang, Kou, Chen, Hsieh, and Gu]{chen2024does}
Chen, Z., Zhang, J., Kou, Y., Chen, X., Hsieh, C.-J., and Gu, Q.
\newblock Why does sharpness-aware minimization generalize better than sgd?
\newblock \emph{Advances in Neural Information Processing Systems}, 36, 2024.

\bibitem[Cimpoi et~al.(2014)Cimpoi, Maji, Kokkinos, Mohamed, , and Vedaldi]{cimpoi14describing}
Cimpoi, M., Maji, S., Kokkinos, I., Mohamed, S., , and Vedaldi, A.
\newblock Describing textures in the wild.
\newblock In \emph{Proceedings of the {IEEE} Conf. on Computer Vision and Pattern Recognition ({CVPR})}, 2014.

\bibitem[Dembo(2009)]{dembo2009large}
Dembo, A.
\newblock \emph{Large deviations techniques and applications}.
\newblock Springer, 2009.

\bibitem[Deng et~al.(2009)Deng, Dong, Socher, Li, Li, and Fei-Fei]{deng2009imagenet}
Deng, J., Dong, W., Socher, R., Li, L.-J., Li, K., and Fei-Fei, L.
\newblock Imagenet: A large-scale hierarchical image database.
\newblock In \emph{2009 IEEE conference on computer vision and pattern recognition}, pp.\  248--255. Ieee, 2009.

\bibitem[Ding et~al.(2024)Ding, Drusvyatskiy, Fazel, and Harchaoui]{ding2024flat}
Ding, L., Drusvyatskiy, D., Fazel, M., and Harchaoui, Z.
\newblock Flat minima generalize for low-rank matrix recovery.
\newblock \emph{Information and Inference: A Journal of the IMA}, 13\penalty0 (2):\penalty0 iaae009, 2024.

\bibitem[Dosovitskiy et~al.(2020)Dosovitskiy, Beyer, Kolesnikov, Weissenborn, Zhai, Unterthiner, Dehghani, Minderer, Heigold, Gelly, et~al.]{dosovitskiy2020image}
Dosovitskiy, A., Beyer, L., Kolesnikov, A., Weissenborn, D., Zhai, X., Unterthiner, T., Dehghani, M., Minderer, M., Heigold, G., Gelly, S., et~al.
\newblock An image is worth 16x16 words: Transformers for image recognition at scale.
\newblock \emph{arXiv preprint arXiv:2010.11929}, 2020.

\bibitem[Du et~al.(2022)Du, Zhou, Feng, Tan, and Zhou]{du2022sharpness}
Du, J., Zhou, D., Feng, J., Tan, V., and Zhou, J.~T.
\newblock Sharpness-aware training for free.
\newblock \emph{Advances in Neural Information Processing Systems}, 35:\penalty0 23439--23451, 2022.

\bibitem[Duchi et~al.(2011)Duchi, Hazan, and Singer]{duchi2011adaptive}
Duchi, J., Hazan, E., and Singer, Y.
\newblock Adaptive subgradient methods for online learning and stochastic optimization.
\newblock \emph{Journal of machine learning research}, 12\penalty0 (7), 2011.

\bibitem[Duchi et~al.(2012)Duchi, Bartlett, and Wainwright]{duchi2012randomized}
Duchi, J.~C., Bartlett, P.~L., and Wainwright, M.~J.
\newblock Randomized smoothing for stochastic optimization.
\newblock \emph{SIAM Journal on Optimization}, 22\penalty0 (2):\penalty0 674--701, 2012.

\bibitem[Foret et~al.(2020)Foret, Kleiner, Mobahi, and Neyshabur]{foret2020sharpness}
Foret, P., Kleiner, A., Mobahi, H., and Neyshabur, B.
\newblock Sharpness-aware minimization for efficiently improving generalization.
\newblock \emph{arXiv preprint arXiv:2010.01412}, 2020.

\bibitem[He et~al.(2016)He, Zhang, Ren, and Sun]{he2016deep}
He, K., Zhang, X., Ren, S., and Sun, J.
\newblock Deep residual learning for image recognition.
\newblock In \emph{Proceedings of the IEEE conference on computer vision and pattern recognition}, pp.\  770--778, 2016.

\bibitem[Johnson \& Zhang(2013)Johnson and Zhang]{johnson2013accelerating}
Johnson, R. and Zhang, T.
\newblock Accelerating stochastic gradient descent using predictive variance reduction.
\newblock \emph{Advances in neural information processing systems}, 26, 2013.

\bibitem[Kingma \& Ba(2014)Kingma and Ba]{kingma2014adam}
Kingma, D.~P. and Ba, J.
\newblock Adam: A method for stochastic optimization.
\newblock \emph{arXiv preprint arXiv:1412.6980}, 2014.

\bibitem[Kort \& Bertsekas(1972)Kort and Bertsekas]{kort1972new}
Kort, B.~W. and Bertsekas, D.~P.
\newblock A new penalty function method for constrained minimization.
\newblock In \emph{Proceedings of the 1972 ieee conference on decision and control and 11th symposium on adaptive processes}, pp.\  162--166. IEEE, 1972.

\bibitem[Krizhevsky et~al.(2009)Krizhevsky, Hinton, et~al.]{krizhevsky2009learning}
Krizhevsky, A., Hinton, G., et~al.
\newblock Learning multiple layers of features from tiny images.
\newblock 2009.

\bibitem[Kwon et~al.(2021)Kwon, Kim, Park, and Choi]{kwon2021asam}
Kwon, J., Kim, J., Park, H., and Choi, I.~K.
\newblock Asam: Adaptive sharpness-aware minimization for scale-invariant learning of deep neural networks.
\newblock In \emph{International Conference on Machine Learning}, pp.\  5905--5914. PMLR, 2021.

\bibitem[Le \& Yang(2015)Le and Yang]{le2015tiny}
Le, Y. and Yang, X.
\newblock Tiny imagenet visual recognition challenge.
\newblock \emph{CS 231N}, 7\penalty0 (7):\penalty0 3, 2015.

\bibitem[Leimkuhler \& Reich(2004)Leimkuhler and Reich]{leimkuhler2004simulating}
Leimkuhler, B. and Reich, S.
\newblock \emph{Simulating hamiltonian dynamics}.
\newblock Number~14. Cambridge university press, 2004.

\bibitem[Li \& Giannakis(2023)Li and Giannakis]{li2024enhancing}
Li, B. and Giannakis, G.
\newblock Enhancing sharpness-aware optimization through variance suppression.
\newblock \emph{Advances in Neural Information Processing Systems}, 36, 2023.

\bibitem[Li et~al.(2023)Li, Beirami, Sanjabi, and Smith]{li2023tilted}
Li, T., Beirami, A., Sanjabi, M., and Smith, V.
\newblock On tilted losses in machine learning: Theory and applications.
\newblock \emph{Journal of Machine Learning Research}, 24\penalty0 (142):\penalty0 1--79, 2023.

\bibitem[Liu \& Theodorou(2019)Liu and Theodorou]{liu2019deep}
Liu, G.-H. and Theodorou, E.~A.
\newblock Deep learning theory review: An optimal control and dynamical systems perspective.
\newblock \emph{arXiv preprint arXiv:1908.10920}, 2019.

\bibitem[Liu et~al.(2022{\natexlab{a}})Liu, Mai, Chen, Hsieh, and You]{liu2022towards}
Liu, Y., Mai, S., Chen, X., Hsieh, C.-J., and You, Y.
\newblock Towards efficient and scalable sharpness-aware minimization.
\newblock In \emph{Proceedings of the IEEE/CVF Conference on Computer Vision and Pattern Recognition}, pp.\  12360--12370, 2022{\natexlab{a}}.

\bibitem[Liu et~al.(2022{\natexlab{b}})Liu, Mai, Cheng, Chen, Hsieh, and You]{liu2022random}
Liu, Y., Mai, S., Cheng, M., Chen, X., Hsieh, C.-J., and You, Y.
\newblock Random sharpness-aware minimization.
\newblock \emph{Advances in Neural Information Processing Systems}, 35:\penalty0 24543--24556, 2022{\natexlab{b}}.

\bibitem[Long \& Bartlett(2023)Long and Bartlett]{long2023sharpness}
Long, P.~M. and Bartlett, P.~L.
\newblock Sharpness-aware minimization and the edge of stability.
\newblock \emph{arXiv preprint arXiv:2309.12488}, 2023.

\bibitem[Mi et~al.(2022)Mi, Shen, Ren, Zhou, Sun, Ji, and Tao]{mi2022make}
Mi, P., Shen, L., Ren, T., Zhou, Y., Sun, X., Ji, R., and Tao, D.
\newblock Make sharpness-aware minimization stronger: A sparsified perturbation approach.
\newblock \emph{Advances in Neural Information Processing Systems}, 35:\penalty0 30950--30962, 2022.

\bibitem[Mueller et~al.(2023)Mueller, Vlaar, Rolnick, and Hein]{mueller2023normalization}
Mueller, M., Vlaar, T.~J., Rolnick, D., and Hein, M.
\newblock Normalization layers are all that sharpness-aware minimization needs.
\newblock In \emph{Thirty-seventh Conference on Neural Information Processing Systems}, 2023.

\bibitem[Neal et~al.(2011)]{neal2011mcmc}
Neal, R.~M. et~al.
\newblock Mcmc using hamiltonian dynamics.
\newblock \emph{Handbook of markov chain monte carlo}, 2\penalty0 (11):\penalty0 2, 2011.

\bibitem[Nesterov()]{nesterov269method}
Nesterov, Y.
\newblock A method of solving a convex programming problem with convergence rate of $1/k^2$.
\newblock \emph{Proceedings of the USSR Academy of Sciences}, 269:\penalty0 3.

\bibitem[Nesterov(1983)]{nesterov1983method}
Nesterov, Y.
\newblock A method for solving the convex programming problem with convergence rate o (1/k2).
\newblock In \emph{Dokl akad nauk Sssr}, volume 269, pp.\  543, 1983.

\bibitem[Nesterov(2013)]{nesterov2013introductory}
Nesterov, Y.
\newblock \emph{Introductory lectures on convex optimization: A basic course}, volume~87.
\newblock Springer Science \& Business Media, 2013.

\bibitem[Rice et~al.(2021)Rice, Bair, Zhang, and Kolter]{rice2021robustness}
Rice, L., Bair, A., Zhang, H., and Kolter, J.~Z.
\newblock Robustness between the worst and average case.
\newblock \emph{Advances in Neural Information Processing Systems}, 34:\penalty0 27840--27851, 2021.

\bibitem[Robey et~al.(2022)Robey, Chamon, Pappas, and Hassani]{robey2022probabilistically}
Robey, A., Chamon, L., Pappas, G.~J., and Hassani, H.
\newblock Probabilistically robust learning: Balancing average and worst-case performance.
\newblock In \emph{International Conference on Machine Learning}, pp.\  18667--18686. PMLR, 2022.

\bibitem[Rockafellar et~al.(2000)Rockafellar, Uryasev, et~al.]{rockafellar2000optimization}
Rockafellar, R.~T., Uryasev, S., et~al.
\newblock Optimization of conditional value-at-risk.
\newblock \emph{Journal of risk}, 2:\penalty0 21--42, 2000.

\bibitem[Sanh(2019)]{sanh2019distilbert}
Sanh, V.
\newblock Distilbert, a distilled version of bert: Smaller, faster, cheaper and lighter.
\newblock \emph{arXiv preprint arXiv:1910.01108}, 2019.

\bibitem[Shalev-Shwartz \& Ben-David(2014)Shalev-Shwartz and Ben-David]{shalev2014understanding}
Shalev-Shwartz, S. and Ben-David, S.
\newblock \emph{Understanding machine learning: From theory to algorithms}.
\newblock Cambridge university press, 2014.

\bibitem[Shen \& Li(2010)Shen and Li]{shen2010dual}
Shen, C. and Li, H.
\newblock On the dual formulation of boosting algorithms.
\newblock \emph{IEEE Transactions on Pattern Analysis and Machine Intelligence}, 32\penalty0 (12):\penalty0 2216--2231, 2010.

\bibitem[Siegmund(1976)]{siegmund1976importance}
Siegmund, D.
\newblock Importance sampling in the monte carlo study of sequential tests.
\newblock \emph{The Annals of Statistics}, pp.\  673--684, 1976.

\bibitem[Streeter \& McMahan(2010)Streeter and McMahan]{streeter2010less}
Streeter, M. and McMahan, H.~B.
\newblock Less regret via online conditioning.
\newblock \emph{arXiv preprint arXiv:1002.4862}, 2010.

\bibitem[Szab{\'o} et~al.(2021)Szab{\'o}, Jamali-Rad, and Mannava]{szabo2021tilted}
Szab{\'o}, A., Jamali-Rad, H., and Mannava, S.-D.
\newblock Tilted cross-entropy (tce): Promoting fairness in semantic segmentation.
\newblock In \emph{Proceedings of the IEEE/CVF Conference on Computer Vision and Pattern Recognition}, pp.\  2305--2310, 2021.

\bibitem[Tahmasebi et~al.(2024)Tahmasebi, Soleymani, Bahri, Jegelka, and Jaillet]{TahmasebiSBJJ24}
Tahmasebi, B., Soleymani, A., Bahri, D., Jegelka, S., and Jaillet, P.
\newblock A universal class of sharpness-aware minimization algorithms.
\newblock In \emph{ICML}, 2024.

\bibitem[Wainwright \& Jordan(2008)Wainwright and Jordan]{wainwright2008graphical}
Wainwright, M.~J. and Jordan, M.~I.
\newblock Graphical models, exponential families, and variational inference.
\newblock \emph{Foundations and Trends{\textregistered} in Machine Learning}, 2008.

\bibitem[Wen et~al.(2022)Wen, Ma, and Li]{wen2022does}
Wen, K., Ma, T., and Li, Z.
\newblock How does sharpness-aware minimization minimize sharpness?
\newblock \emph{arXiv preprint arXiv:2211.05729}, 2022.

\bibitem[Wen et~al.(2024)Wen, Li, and Ma]{wen2024sharpness}
Wen, K., Li, Z., and Ma, T.
\newblock Sharpness minimization algorithms do not only minimize sharpness to achieve better generalization.
\newblock \emph{Advances in Neural Information Processing Systems}, 36, 2024.

\bibitem[Whittle(1981)]{whittle1981risk}
Whittle, P.
\newblock Risk-sensitive linear/quadratic/gaussian control.
\newblock \emph{Advances in Applied Probability}, 13\penalty0 (4):\penalty0 764--777, 1981.

\bibitem[Whittle(2002)]{whittle2002risk}
Whittle, P.
\newblock Risk sensitivity, a strangely pervasive concept.
\newblock \emph{Macroeconomic Dynamics}, 6\penalty0 (1):\penalty0 5--18, 2002.

\bibitem[Xie et~al.(2024)Xie, Latorre, Antonakopoulos, Pethick, and Cevher]{pmlr-v235-xie24d}
Xie, W., Latorre, F., Antonakopoulos, K., Pethick, T., and Cevher, V.
\newblock Improving {SAM} requires rethinking its optimization formulation.
\newblock In \emph{Proceedings of the 41st International Conference on Machine Learning}, 2024.

\bibitem[Zagoruyko(2016)]{zagoruyko2016wide}
Zagoruyko, S.
\newblock Wide residual networks.
\newblock \emph{arXiv preprint arXiv:1605.07146}, 2016.

\bibitem[Zhang et~al.(2024)Zhang, Li, and Ju]{zhang2024noise}
Zhang, H.~R., Li, D., and Ju, H.
\newblock Noise stability optimization for finding flat minima: A hessian-based regularization approach.
\newblock \emph{Transactions on Machine Learning Research}, 2024.
\newblock ISSN 2835-8856.

\bibitem[Zhao et~al.(2022)Zhao, Zhang, and Hu]{zhao2022penalizing}
Zhao, Y., Zhang, H., and Hu, X.
\newblock Penalizing gradient norm for efficiently improving generalization in deep learning.
\newblock In \emph{International Conference on Machine Learning}, pp.\  26982--26992. PMLR, 2022.

\bibitem[Zheng et~al.(2021)Zheng, Zhang, and Mao]{zheng2021regularizing}
Zheng, Y., Zhang, R., and Mao, Y.
\newblock Regularizing neural networks via adversarial model perturbation.
\newblock In \emph{Proceedings of the IEEE/CVF Conference on Computer Vision and Pattern Recognition}, pp.\  8156--8165, 2021.

\bibitem[Zhou et~al.(2020)Zhou, Wang, and Bilmes]{zhou2020robust}
Zhou, T., Wang, S., and Bilmes, J.
\newblock Robust curriculum learning: from clean label detection to noisy label self-correction.
\newblock In \emph{International Conference on Learning Representations}, 2020.

\bibitem[Zhou et~al.(2021)Zhou, Liu, Zhang, and Chen]{zhou2021sharpness}
Zhou, W., Liu, F., Zhang, H., and Chen, M.
\newblock Sharpness-aware minimization with dynamic reweighting.
\newblock \emph{arXiv preprint arXiv:2112.08772}, 2021.

\bibitem[Zhuang et~al.(2022)Zhuang, Gong, Yuan, Cui, Adam, Dvornek, Tatikonda, Duncan, and Liu]{zhuang2022surrogate}
Zhuang, J., Gong, B., Yuan, L., Cui, Y., Adam, H., Dvornek, N., Tatikonda, S., Duncan, J., and Liu, T.
\newblock Surrogate gap minimization improves sharpness-aware training.
\newblock \emph{arXiv preprint arXiv:2203.08065}, 2022.

\end{thebibliography}
\bibliographystyle{icml2025}

\newpage
\appendix
\onecolumn

\section{Additional Toy Problems} \label{app:toy}

In Figure~\ref{fig:toy} in Section~\ref{fig:toy}, we present a specific toy problem where TSAM arrives at more flat solutions as $t$ increases. Though the TSAM objective will recover SAM when $t\to \infty$, we note that TSAM can be easier to solve due to smoothness. To illustrate this, we create another toy problem in Figure~\ref{fig:toy2-sam} and~\ref{fig:toy2-tsam} below. We see that SAM always leads to a non-smooth optimization problem for $\rho>0$. 

\begin{figure}[h!]
    \centering
    \includegraphics[width=1\linewidth]{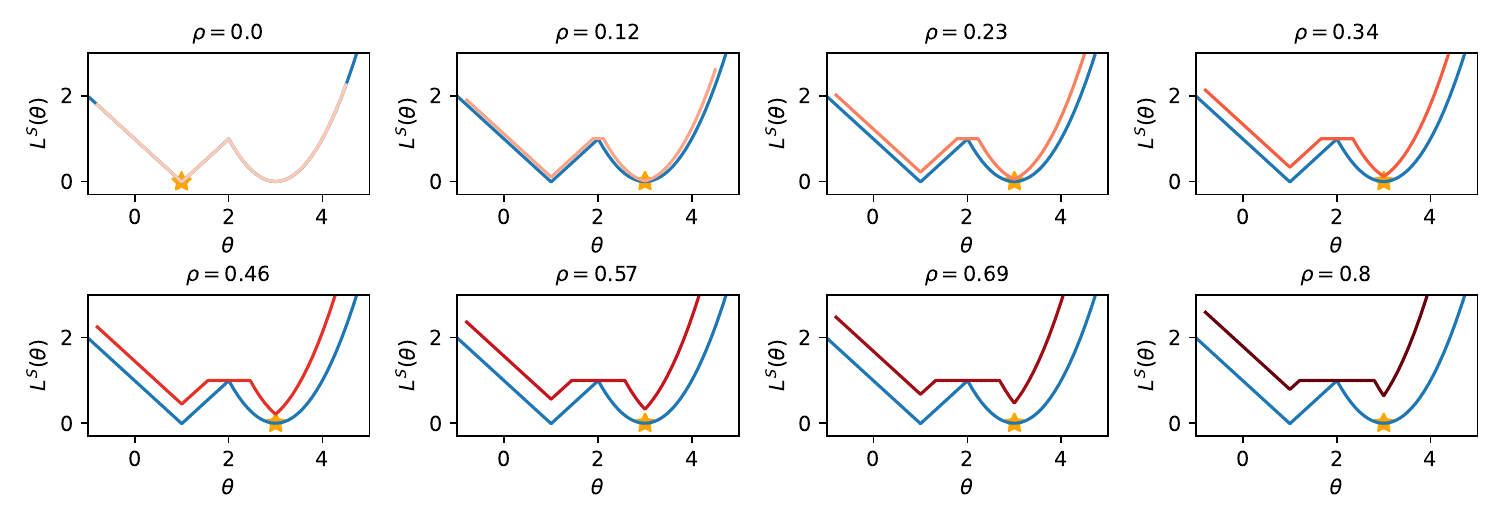}
    \caption{\textbf{SAM losses as $\rho$ increases.} The original loss function (shown in the \textcolor{blue}{blue} lines across all figures) is a one-dimensional problem $L(\theta) = |\theta-1|-0.01 \textit{\text{~if~}} \theta \leq 2$, and $L(\theta)=(\theta-3)^2$ \textit{otherwise}. Note that $\theta=3$ is a more flat solution than $\theta=1$, though $L(1) < L(3)$.
    The SAM objective is $\min_{\theta} \max_{|\epsilon| \leq \rho} L(\theta+\epsilon)$, shown in the \textcolor{red}{red} lines, where the values of $\rho$'s increase from 0 to 0.8. When $\rho=0$, the objective reduces to ERM. For $\rho>0$, the SAM objectives (red lines) are non-smooth, and the global minima (marked in \textcolor{orange}{orange}) are achieved at a flat region in $L(\cdot)$. The SAM objective visualization holds \textit{regardless of the usage of any existing SAM algorithms or implementation.}}
    \label{fig:toy2-sam}
\end{figure}

\begin{figure}[h!]
    \centering
    \includegraphics[width=1\linewidth]{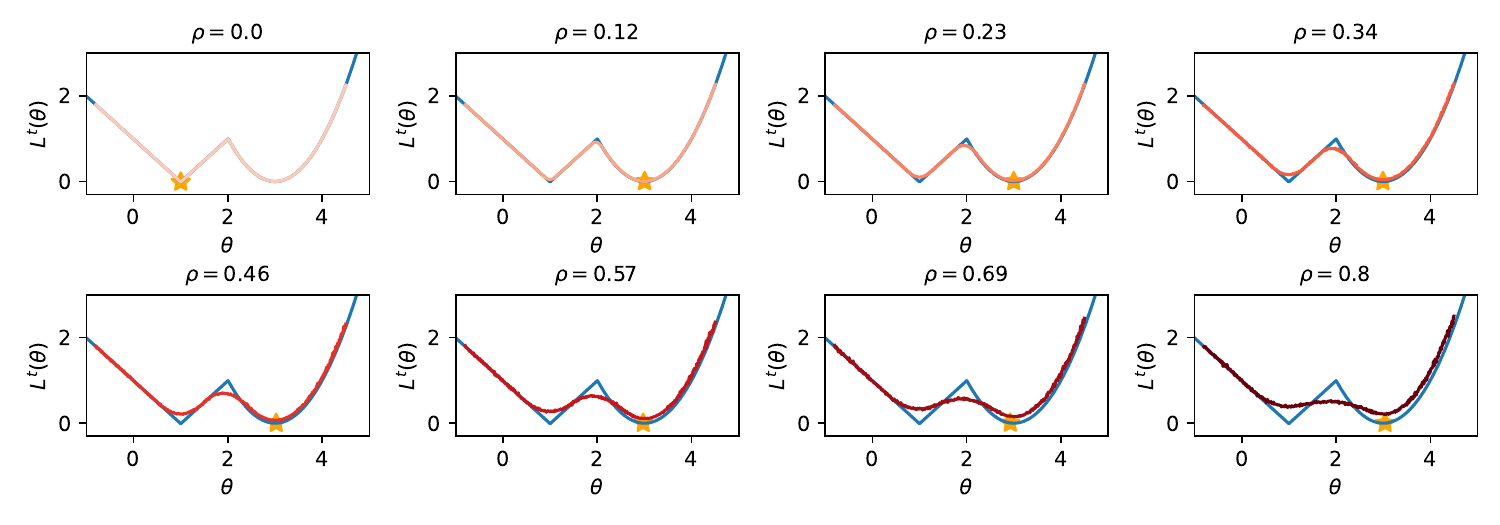}
    \caption{ \textbf{TSAM ($t$=0.01) losses as $\rho$ increases.} The TSAM objective (\textcolor{red}{red} lines) is $\frac{1}{t} \log \left(\mathbb{E}_{\mu(\epsilon)} \left[e^{t L(\theta+\epsilon)}\right]\right)$, where $\mu(\epsilon) := \mathcal{U}(|\epsilon| \leq \rho)$ defines a uniform distribution of $\epsilon$'s constrained in a ball with radius $\rho$. The larger $\rho$ is, the darker the redness becomes. TSAM with a small $t$ is able to find flat solutions.}
    \label{fig:toy2-tsam}
\end{figure}

\newpage
\section{Complete Proofs} \label{app:proof:properties}

\subsection{Proofs for Section~\ref{sec:properties:general}}

\paragraph{Proof for the Case of $t \to 0$, $L^t(\theta+\epsilon) \to 
\mathbb{E}[L(\theta+\epsilon)]$.}
Note that if $L(\cdot)$ is continuously differentiable, then $e^{tL(\theta+\epsilon)}$ is continuous w.r.t. $\epsilon \in \mathbb{R}^d$. It is also continuous w.r.t. $t \in \mathbb{R}$. When $t \to 0$, 
\begin{align}
    \lim_{t \to 0} L^t(\theta) &= \lim_{t \to 0} \frac{1}{t} \log \left(\int e^{tL(\theta+\epsilon)} d \mu(\epsilon) \right) \\&= \frac{1}{\int e^{tL(\theta+\epsilon)} d \mu(\epsilon) } \int e^{t L(\theta+\epsilon)} L(\theta+\epsilon) d \mu(\epsilon) \\&= \int L(\theta+\epsilon) d \mu(\epsilon) \\&= \mathbb{E}[L(\theta+\epsilon)].
\end{align}

\paragraph{Proof for Lipschitzness.} First observe that if $L(\theta)$ is $p$-Lipschitz with respect to $\theta$, then $L^t(\theta)$ is $p$-Lipschitz with respect to $\theta$. This follows from
\begin{align}
    \left |L^t(\theta_1) -L^t(\theta_2) \right | &=  \left |\frac{1}{t}\log \mathbb{E}\left[e^{t L(\theta_1+\epsilon)}\right] -  \frac{1}{t}\log \mathbb{E}\left[e^{t L(\theta_2+\epsilon)}\right] \right | \\
    &= \left |\frac{1}{t}\log \frac{\mathbb{E}\left[e^{t L(\theta_1+\epsilon)}\right]}{\mathbb{E}\left[e^{t L(\theta_2+\epsilon)}\right]} \right | \\
    &\leq  \left |\frac{1}{t}\log e^{t p \|\theta_1-\theta_2\|} \frac{\mathbb{E}\left[e^{t L(\theta_2+\epsilon)}\right]}{\mathbb{E}\left[e^{t L(\theta_2+\epsilon)}\right]} \right | \\
    &=p \|\theta_1-\theta_2\|.
\end{align}

\paragraph{Proof for Strong Convexity.} Assume $L$ is continuously differentiable. If $L$ is $\mu$-strongly convex, then $L^t$ is also $\mu$-strongly convex. This is because of the Hessian in Eq.~(\ref{eq:hessian}), which can be written as
\begin{align}
    \nabla^2 L^t(\theta) = t \cdot M + \mathbb{E}\left[\frac{e^{t L(\theta+\epsilon)}}{\mathbb{E}[e^{tL(\theta+\epsilon)}]} \nabla^2 L(\theta+\epsilon)\right], \label{eq:hessian}
\end{align}
where $M$ is a positive semi-definite matrix. We note that due to the $\mu$-strong convexity of $L$, the second term satisfies $\mathbb{E}\left[\frac{e^{t L(\theta+\epsilon)}}{\mathbb{E}[e^{tL(\theta+\epsilon)}]} \nabla^2 L(\theta+\epsilon)\right] \succcurlyeq \mu \mathbf{I}$. Hence, $\nabla^2 L^t(\theta) \succcurlyeq \mu \mathbf{I}$.

\paragraph{Proof for Smoothness.} From Eq.~(\ref{eq:hessian}), we know that
\begin{align}
    \frac{1}{t}\nabla^2 L^t(\theta) = M + \frac{1}{t}\mathbb{E}\left[\frac{e^{t L(\theta+\epsilon)}}{\mathbb{E}[e^{tL(\theta+\epsilon)}]} \nabla^2 L(\theta+\epsilon)\right].
\end{align}
As $\mathbb{E}\left[\frac{e^{t L(\theta+\epsilon)}}{\mathbb{E}[e^{tL(\theta+\epsilon)}]}\right]=1$, and the max eigenvalue $\lambda_{\max}(\nabla^2 L(\theta+\epsilon)) \leq \beta$, we have
\begin{align}
    0 < \min_{t \to\infty} \frac{1}{t} \lambda_{\max}(\nabla^2 L(\theta+\epsilon)) < + \infty.
\end{align}

\subsection{Proofs for Section~\ref{sec:properties:glms}}

In the following, we use $\mathbb{E}$ to denote $\mathbb{E}_{\epsilon}$.
Define $g(t)$ as
\begin{align}
    g(t) &:= L^t(\theta_1) - L^t(\theta_2) \\
    &= \frac{1}{t} \log \left(\int e^{tL(\theta_1+\epsilon)} d \mu(\epsilon) \right) - \frac{1}{t} \log \left(\int e^{tL(\theta_2+\epsilon)} d \mu(\epsilon) \right). 
\end{align}

Assume $l$ has the specific form of 
\begin{align}
l(x_i;\theta) &= A(\theta)-\theta^\top T(x_i),  \\
    L(\theta) &= A(\theta)-\theta^\top \left(\frac{1}{n}\sum_{i=1}^n T(x_i)\right) := A(\theta)-\theta^\top  T(x).
\end{align} 
Under this form, we have that 
\begin{align}
    L^t(\theta) &= \frac{1}{t}\log \left(\int e^{t (A(\theta+\epsilon)-(\theta+\epsilon)^\top T(x))} p(\epsilon) d \epsilon \right) \\
    &= \frac{1}{t}\log \left( e^{-t \theta^\top T(x)}\int e^{t (A(\theta+\epsilon)-\epsilon^\top T(x))} p(\epsilon) d \epsilon \right) \\
    &= -\theta^\top T(x) + \frac{1}{t} \log \left(\int e^{t (A(\theta+\epsilon)-\epsilon^\top T(x))} p(\epsilon) d\epsilon \right)
\end{align}

Define
\begin{align}
    \Gamma^t(\theta) := \log \left(\mathbb{E}\left[e^{t A(\theta+\epsilon)-t \epsilon^\top T(x)}\right]\right).
\end{align}
Then we have
\begin{align}
    L^t(\theta) = \frac{1}{t}\log\left(\mathbb{E}\left[e^{t \left(A(\theta+\epsilon)-\left(\theta+\epsilon\right)^\top T(x)\right)}\right]\right)  = -\theta^\top T(x) + \frac{1}{t} \Gamma^t(\theta).
\end{align} 
Define
\begin{align}
    n^t(\theta) &:= e^{tA(\theta+\epsilon) - t \epsilon^\top T(x)}, \\
    h^t(\theta) &:= \frac{\mathbb{E}\left[n^t(\theta) (A(\theta+\epsilon)-\epsilon^\top T(x)\right]}{\mathbb{E}\left[n^t(\theta)\right]}, \\
    m^t(\theta) &:= -\frac{1}{t^2} \Gamma^t(\theta)  + \frac{1}{t} h^t(\theta). 
\end{align}
We have that
\begin{align}
    \frac{\partial n^t(\theta)}{\partial t} &= n^t(\theta) (A(\theta+\epsilon)-\epsilon^\top T(x)).
\end{align}
We know
\begin{align}
    \frac{\partial \Gamma^t (\theta)}{\partial t} &= \frac{\mathbb{E}\left[e^{t A(\theta+\epsilon)-t \epsilon^\top T(x)} (A(\theta+\epsilon)-\epsilon^\top T(x))\right]}{\mathbb{E}\left[e^{t A(\theta+\epsilon)-t \epsilon^\top T(x)}\right]} =h^t(\theta), \\
    \frac{\partial L^t(\theta)}{\partial t} &= -\frac{1}{t^2} \Gamma^t(\theta) + \frac{1}{t} \frac{\partial \Gamma^t(\theta)}{\partial t} =  -\frac{1}{t^2} \Gamma^t(\theta)  + \frac{1}{t} h^t(\theta) = m^t(\theta).
\end{align}
\subsection{$L^t(\theta)$ is monotonically non-increasing as $t$}
We would like to prove the sign of $\frac{\partial L^t(\theta)}{\partial t}$, or $m^t(\theta)$, is non-negative. The sign of $m^t(\theta)$ is the same as the sign of $t^2 m^t(\theta)$. We have
\begin{align}
    t^2 m^t(\theta) &= -\Gamma^t(\theta) + t h^t(\theta), \\
    \frac{\partial (t^2 m^t(\theta))}{\partial t} &= -h^t(\theta) + h^t(\theta) + t \frac{\partial h^t(\theta)}{\partial t} = t \frac{\partial h^t(\theta)}{\partial t},
\end{align}
and
\begin{align}
    \frac{\partial h^t(\theta)}{\partial t} &= \frac{\mathbb{E}[n^t(\theta) (A(\theta+\epsilon)-\epsilon^\top T(x))^2] \mathbb{E}[n^t(\theta)]}{\left(\mathbb{E}\left[n^t(\theta)\right]\right)^2} \\ &~ - \frac{\mathbb{E}[n^t(\theta) (A(\theta+\epsilon)-\epsilon^\top T(x))] \mathbb{E}[n^t(\theta) (A(\theta+\epsilon)-\epsilon^\top T(x))]}{\left(\mathbb{E}\left[n^t(\theta)\right]\right)^2}.
\end{align}
Let random variables $X = \sqrt{n^t(\theta)} (A(\theta+\epsilon)-\epsilon^\top T(x))$, and $Y=\sqrt{n^t(\theta)}$.  Following the fact $\mathbb{E}[X^2]\mathbb{E}[Y^2] - (\mathbb{E}[XY])^2 \geq 0$ gives
\begin{align}
   \frac{\partial h^t(\theta)}{\partial t} = \frac{\mathbb{E}[X^2]\mathbb{E}[Y^2]-\mathbb{E}[XY] \mathbb{E}[XY]}{\left(\mathbb{E}\left[n^t(\theta)\right]\right)^2} \geq 0.
\end{align}
Therefore,
\begin{align}
  \frac{\partial (t^2 m^t(\theta))}{\partial t} \geq 0.
\end{align}
We note that $\lim_{t \to 0} t^2 m^t(\theta) = \lim_{t \to 0} t^2 -\Gamma^t(\theta) + t h^t(\theta)=0$. Hence $t^2 m^t(\theta) \geq 0$. Therefore, $m^t(\theta) \geq 0$. we have shown that the tilted SAM loss $L^t(\theta)$ is monotonically non-decreasing as the increase of $t$, for any $\theta$. 
\subsection{$t$-SAM prefers  flatter models as $t$ increases}
Next, we examine $g^t(\theta_1, \theta_2) := L^t(\theta_1)-L^t(\theta_2)$. First, we have
\begin{align}
    \frac{\partial g^t(\theta_1, \theta_2)}{\partial t} &= \frac{\partial L^t(\theta_1)}{\partial t}-\frac{\partial L^t(\theta_2)} {\partial t} \\
    &= -\frac{1}{t^2} \Gamma^t(\theta_1) + \frac{1}{t} h^t(\theta_1) + \frac{1}{t^2} \Gamma^t(\theta_2) - \frac{1}{t} h^t(\theta_2) \\
    &= m^t(\theta_1) - m^t(\theta_2). 
\end{align}
Similarly, it holds that
\begin{align}
     \frac{\partial (t^2 m^t(\theta_1))}{\partial t} -  \frac{\partial (t^2 m^t(\theta_2))}{\partial t}  = t \frac{\partial h^t(\theta_1)}{\partial t} - t \frac{\partial h^t(\theta_2)}{\partial t}.
\end{align}
For $t \geq 0$, we have
\begin{align}
    \text{sign} \left(\frac{\partial h^t(\theta_1)}{\partial t}-\frac{\partial h^t (\theta_2)}{\partial t}\right) &= \text{sign} \left(\frac{\partial (t^2 m^t (\theta_1))}{\partial t} - \frac{\partial (t^2 m^t (\theta_2))}{\partial t}\right), \\
    \text{sign} \left(m^t(\theta_1)-m^t(\theta_2)\right) &= \text{sign} \left(t^2 m^t(\theta_1)-t^2 m^t (\theta_2)\right).
\end{align}
Let the random variable $L_1$ denote $A(\theta_1+\epsilon) -\epsilon^\top T(x)$, and random variable $L_2$ denote $A(\theta_2+\epsilon) -\epsilon^\top T(x)$. Then
\begin{align}
    &\frac{\partial h^t(\theta_1)}{\partial t}  = \frac{\mathbb{E}\left[e^{tL_1} L_1^2\right] \mathbb{E}\left[e^{t L_1}\right] - \left(\mathbb{E}\left[e^{tL_1} L_1\right]\right)^2} {\left(\mathbb{E}\left[e^{t L_1}\right]\right)^2} = \frac{\mathbb{E}\left[e^{tL_1} L_1^2\right]}{\mathbb{E}\left[e^{tL_1}\right]} - \left(\frac{\mathbb{E}[e^{tL_1} L_1]}{\mathbb{E}[e^{tL_1}]}\right)^2, \\
    &\frac{\partial h^t(\theta_1)}{\partial t} -  \frac{\partial h^t(\theta_2)}{\partial t} = \frac{\mathbb{E}\left[e^{tL_1} L_1^2\right]}{\mathbb{E}\left[e^{tL_1}\right]} - \left(\frac{\mathbb{E}[e^{tL_1} L_1]}{\mathbb{E}[e^{tL_1}]}\right)^2 - \left(\frac{\mathbb{E}\left[e^{tL_2} L_2^2\right]}{\mathbb{E}\left[e^{tL_2}\right]} - \left(\frac{\mathbb{E}[e^{tL_2} L_2]}{\mathbb{E}[e^{tL_2}]}\right)^2 \right).
\end{align}
Given random variables $L_1$ and $L_2$, the \textit{exponentially reweighted} losses can be defined as $e^{tL_1} L_1$ and $e^{tL_2} L_2$. The $t$-\textit{weighted} second moment is $\frac{\mathbb{E}\left[e^{tL_1} L_1^2\right]}{\mathbb{E}\left[e^{tL_1}\right]}$, and the $t$-\textit{weighted}  mean is $\frac{\mathbb{E}[e^{tL_1} L_1]}{\mathbb{E}[e^{tL_1}]}$. Hence, $\frac{\mathbb{E}\left[e^{tL_1} L_1^2\right]}{\mathbb{E}\left[e^{tL_1}\right]} - \left(\frac{\mathbb{E}[e^{tL_1} L_1]}{\mathbb{E}[e^{tL_1}]}\right)^2$ can be viewed as $t$-\textit{weighted} variance. As $\theta_1$ is $t$-sharper than $\theta_2$, we have $\frac{\partial h^t(\theta_1)}{\partial t} -  \frac{\partial h^t(\theta_2)}{\partial t} \geq 0$. Therefore $t^2 m^t(\theta_1) - t^2 m^t(\theta_2)$ is non-decreasing as $t$ increases. It takes value of $0$ when $t=0$, which implies that $t^2 m^t(\theta_1) - t^2 m^t(\theta_2) = \frac{\partial g^t(\theta_1, \theta_2)}{\partial t}\geq 0$.

\paragraph{Proof for the Discussions on $t \to 0$.}
Recall that  $\exp$  and $\log$ functions can be expanded as 
\begin{align}
    \exp(x) &= 1+\sum_{k=1}^{\infty} \frac{x^{k}}{k!} \approx 1+x+\frac{1}{2} x^2, \\
    \log(x+1) &= \sum_{k=1}^{\infty} (-1)^{k-1} \frac{x^k}{k!} \approx x -\frac{x^2}{2} + \frac{x^3}{6}.
\end{align}
For very small $t \leq 1$, 
\begin{align}
    &\frac{1}{t}\log \left(\mathbb{E}\left[e^{t L(\theta+\epsilon)}\right]\right) 	\\ &\approx \frac{1}{t}\log \left(\mathbb{E}\left[1+t L(\theta+\epsilon) + \frac{t^2}{2} L^2(\theta+\epsilon) \right]\right) \\
    & \approx \frac{1}{t} \left(\mathbb{E}\left[tL(\theta+\epsilon) + \frac{t^2}{2} L^2(\theta+\epsilon)\right] - \frac{1}{2} \mathbb{E}^2\left[tL(\theta+\epsilon) + \frac{t^2}{2} L^2(\theta+\epsilon)\right]\right) \\
    & = \frac{1}{t} \left(t\mathbb{E}[ L(\theta+\epsilon)] + \frac{t^2}{2} \mathbb{E}\left[L^2(\theta+\epsilon)\right] - \frac{t^2}{2} \mathbb{E}^2[L(\theta+\epsilon)] +O(t^3)+O(t^4)\right) \\
    &= \mathbb{E}[L(\theta+\epsilon)] + \frac{t}{2} \left(\mathbb{E}\left[L^2(\theta+\epsilon)\right] - \mathbb{E}^2\left[L(\theta+\epsilon)\right]\right) + O(t^2) \\
    &\approx \mathbb{E}[L(\theta+\epsilon)] + \frac{t}{2} \var\left(L(\theta+\epsilon)\right)
\end{align}
Hence, our proposed objective can be viewed as optimizing for the mean plus variance of the losses in the neighborhood regions when $t$ is very close to 0. When $t=0$, it reduces to only optimizing for $\mathbb{E}[L(\theta+\epsilon)]$ for $\epsilon \in \mu (\epsilon)$. 
For any $\theta_1$ and $\theta_2$ such that $\theta_1$ is sharper than $\theta_2$, we have
\begin{align}
    g(t) &\approx \mathbb{E}[L(\theta_1+\epsilon)] + \frac{t}{2} \var \left(L(\theta_1+\epsilon)\right) - \mathbb{E}[L(\theta_2+\epsilon)] - \frac{t}{2} \var \left(L(\theta_2+\epsilon)\right),
    \\ g'(t) &\approx \frac{1}{2} \left(\var \left(L(\theta_1+\epsilon)\right) - \var \left(L(\theta_2+\epsilon)\right)\right).
\end{align}
Recall that sharpness is defined as standard variance when $t \to 0$ (Definition~\ref{def:sharpness}), and we have that $g'(t)\geq 0$.

\subsection{Proof for Theorem~\ref{thm:generalization}}

We first state some useful lemmas.
\begin{lemma}[Stated and proved in~\citet{term-generalization}] \label{lemma:1}
    Let $X$ be a random variable. Suppose $0 < a < X < b < +\infty$, we have
\begin{align}
     \frac{\var(X)}{2b^2} \leq \log(\mathbb{E}[X]) - \mathbb{E}[\log(X)] \leq \frac{\var(X)}{2a^2}.
\end{align}
\end{lemma}

The Lemma directly follows from existing results in~\citet{term-generalization}. For completeness, we include the proof here.
\begin{proof}
    As $\frac{d^2}{d x^2}\left(\log(x) + \beta x^2\right) = \frac{-1}{x^2} + 2\beta,$ the function $\log(x) + \beta x^2$ is concave  for $\beta=\frac{1}{2b^2}$ and convex for $\beta=\frac{1}{2a^2}$.
Hence, by Jensen's inequality,  
\begin{align}
    \mathbb{E}[\log(X)]
    =
     \mathbb{E}\Big[\log(X)+\frac{X^2}{2b^2}-\frac{X^2}{2b^2}\Big] 
    &\leq \log( \mathbb{E}[X])+\frac{1}{2b^2} \mathbb{E}[X]^2-\frac{1}{2b^2} \mathbb{E}[X^2]\\
    &=\log( \mathbb{E}[X])-\frac{1}{2b^2}\var(X),
\end{align}
 which completes the proof of the lower bound. A similar approach can be applied to derive the upper bound.
\end{proof}

\paragraph{Proof for Theorem~\ref{thm:generalization}.} We can now proceed with the detailed proof below.
\begin{proof}
Examine the following decomposition of the generalization error:
\begin{align}
    &\mathbb{E}_{Z} [l(\theta, Z)] - \frac{1}{t} \log \mathbb{E}_{\epsilon} [e^{t L(\theta+\epsilon)}] = \mathbb{E}_{Z} [l(\theta, Z)] - \mathbb{E}_{\epsilon} [L(\theta+\epsilon)]  + \mathbb{E}_{\epsilon} [L(\theta+\epsilon)] - \frac{1}{t} \log \mathbb{E}_{\epsilon} [e^{t L(\theta+\epsilon)}]. 
\end{align}

Based on Lemma~\ref{lemma:1}, let $X$ be $e^{tL(\theta+\epsilon)}$ and $1 \leq e^{tL(\theta+\epsilon)} \leq e^{tM}$ (assuming positive and bounded losses and non-negative $t$), we have that 
\begin{align}
   \frac{\var(e^{tL(\theta+\epsilon)})}{2t e^{2tM}} \leq \frac{1}{t}\log(\mathbb{E}_{\epsilon}[e^{tL(\theta+\epsilon)}]) - \mathbb{E}[ L(\theta+\epsilon)] \leq \frac{\var(e^{t L(\theta+\epsilon)})}{2t},
\end{align}
and 
\begin{align}
   -\frac{\var(e^{tL(\theta+\epsilon)})}{2t e^{2tM}} \geq  \mathbb{E}_{\epsilon}[ L(\theta+\epsilon)] - \frac{1}{t}\log(\mathbb{E}_{\epsilon}[e^{tL(\theta+\epsilon)}]) \geq -\frac{\var(e^{t L(\theta+\epsilon)})}{2t}. \label{eq:term1}
\end{align}
For the term $\mathbb{E}_Z[l(\theta, Z)] - \mathbb{E}_{\epsilon}[L(\theta+\epsilon)]$, we further decompose it into
\begin{align}
    \mathbb{E}_Z[l(\theta, Z)] - \mathbb{E}_{\epsilon}[L(\theta+\epsilon)] =  \mathbb{E}_Z[l(\theta, Z)] - L(\theta) + L(\theta) - \mathbb{E}_\epsilon[L(\theta+\epsilon)]. \label{eq:term2}
\end{align}
Recall that $L(\cdot)$ denote the empirical average loss based on $n$ training samples (Eq.~(\ref{obj:erm})), applying Hoeffding Inequality~\citep{10.1093/acprof:oso/9780199535255.001.0001} gives
\begin{align}
    \mathbb{E}_Z[l(\theta, Z)] - L(\theta) \leq M \sqrt{\frac{\log(2/\delta)}{2n}}. \label{eq:term3}
\end{align}
Combining Eq.(\ref{eq:term1}), (\ref{eq:term2}), and (\ref{eq:term3}), we have the desired bound
\begin{align}
    &\mathbb{E}_{Z} [l(\theta, Z)] - \frac{1}{t} \log \mathbb{E}_{\epsilon} [e^{t L(\theta+\epsilon)}] \leq M \sqrt{\frac{\log(2/\delta)}{2n}} -\frac{\var(e^{tL(\theta+\epsilon)})}{2t e^{2tM}} + L(\theta) - \mathbb{E}_{\epsilon}[L(\theta+\epsilon)].
\end{align}
To investigate the impacts of $t$ on the generalization bound, we can leave the last term $L(\theta) - \mathbb{E}_{\epsilon}[L(\theta+\epsilon)]$ as it is since it is independent of $t$.

Additionally, to bound $L(\theta)-\mathbb{E}_{\epsilon} [L(\theta+\epsilon)]$ (the gap between empirical average losses and its randomly-smoothed version) we note that it is related to the curvature of $L(\theta)$. If we further assume that the loss is $\mu$-strongly convex, then it holds that
\begin{align}
    L(\theta) - \mathbb{E}[L(\theta+\epsilon)] \leq L(\theta) - L(\theta) - \mathbb{E}[\nabla L(\theta)^{\top} \epsilon] -\frac{\mu}{2} \|\epsilon\|^2 \leq -\frac{\mu}{2} \mathbb{E}[\|\epsilon\|^2].
\end{align}
Combining all the above results gives
\begin{align}
    \mathbb{E}_{Z} [l(\theta, Z)] - \frac{1}{t} \log \mathbb{E}_{\epsilon} [e^{t L(\theta+\epsilon)}] \leq M \sqrt{\frac{\log(2/\delta)}{2n}} -\frac{\var(e^{tL(\theta+\epsilon)})}{2t e^{2tM}} - \frac{\mu}{2} \mathbb{E}[\|\epsilon\|^2].
\end{align}
\end{proof}

\newpage
\section{Additional Experimental Details} \label{app:experiments}


\subsection{Hyperparamter Tuning}
For all the three datasets and all methods, we tune learning rates from $\{0.0001, 0.0003, 0.001, 0.003, 0.01, 0.03, 0.1\}$. We use a fixed batch size of 64, label smoothing of 0.1 (for smooth cross-entropy loss), momentum with a parameter 0.9, and $L_2$ weight decay 0.0005 for all runs. For vanilla SAM, we tune $\rho$ from $\{0.05, 0.1, 0.5, 1, 5, 15\}$ and found that the best values are ${0.05, 0.1, 0.1}$ for CIFAR100, DTD, noisy CIFAR100, respectively. For SAM variants, we tune $\rho$ parameters in the same way. The PGN baseline~\citep{zhao2022penalizing} introduces another hyperparameter---the coefficients for the linear combination between normal gradients and SAM gradients, and we tune that from $\{0.3, 0.5, 0.7, 0.9\}$. We follow the recommendations of $\lambda$ and $\gamma$ hyperparameters in the original Random SAM paper~\citep{liu2022random}. For TSAM, we set $\delta=\frac{t}{2}$ in Algorithm~\ref{alg:sampling}, set $\rho$ to be 20, and $\alpha$ in Algorithm~\ref{alg:main} to be 0.995 across all datasets. We tune $t$ from $\{0, 1, 5, 20, 100\}$, and the best $t$'s are 20, 5, 1 for the three image datasets.  The number of sampled $\epsilon$'s (the $s$ hyperparameter in Algorithm~\ref{alg:sampling}) are chosen from $\{3, 5\}$. We show the effects of $t$ and $s$ in detail in Section~\ref{sec:experiments:hyperparameter} in the main text.

\subsection{Naive Sampling}

As discussed in Section~\ref{sec:method}, one naive approach to estimate $\frac{\mathbb{E}[e^{tL(\theta+\epsilon)} \nabla L(\theta+\epsilon)]}{\mathbb{E}[e^{tL(\theta+\epsilon)}]}$ for $\epsilon$ in uniformly distributed over $\mu(\epsilon)$ is to first uniformly sample $s$ $\epsilon$'s over $\mu(\epsilon)$, and then perform tilted aggregation, as follows:
\begin{align}
    \hat{g} \leftarrow \frac{\sum_{i \in [s]} e^{tL(\theta+\epsilon_i)} \nabla L(\theta+\epsilon_i)}{\sum_{i \in [s]} e^{L(\theta+\epsilon_i)}}, ~\epsilon_i \sim \mu(\epsilon).
\end{align}
We demonstrate convergence as a function of $s$ on the CIFAR100 dataset in the figure below. We see that as $s$ increases, the performance increases. However, when $s=10$, which means that we need 10 gradient evaluations per model updates, the accuracy is lower than that of TSAM with the proposed algorithm.

\begin{figure}[h!]
    \centering
    \includegraphics[width=0.36\textwidth]{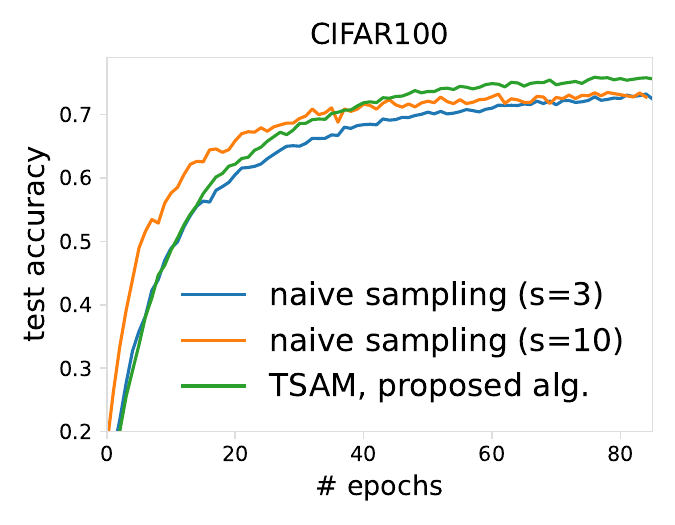}
    \vspace{-1em}
    \caption{TSAM ($t=20$) under the proposed algorithm compared with first uniformly sampling $\epsilon$ and then performing tilted aggregation.}
    \label{fig:compare_random_sampling}
\end{figure}

\subsection{Additional Results}

\begin{figure}[h!]
    \centering
    \begin{subfigure}{0.32\textwidth}
        \centering
        \includegraphics[width=\textwidth]{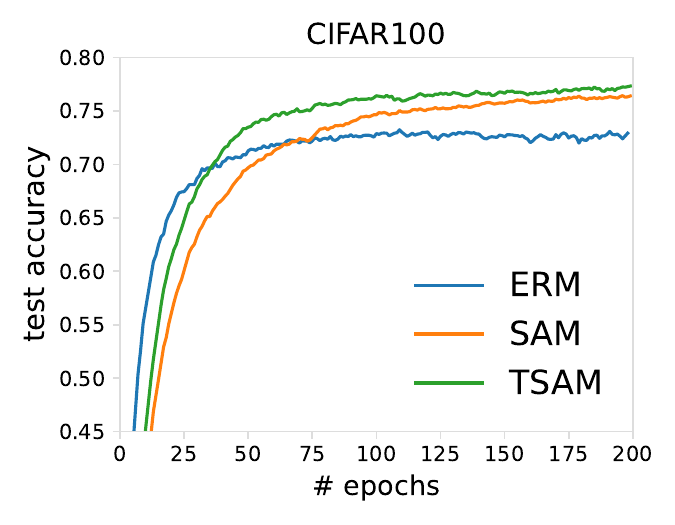}
    \end{subfigure}
    \hfill
    \begin{subfigure}{0.32\textwidth}
        \centering
        \includegraphics[width=\textwidth]{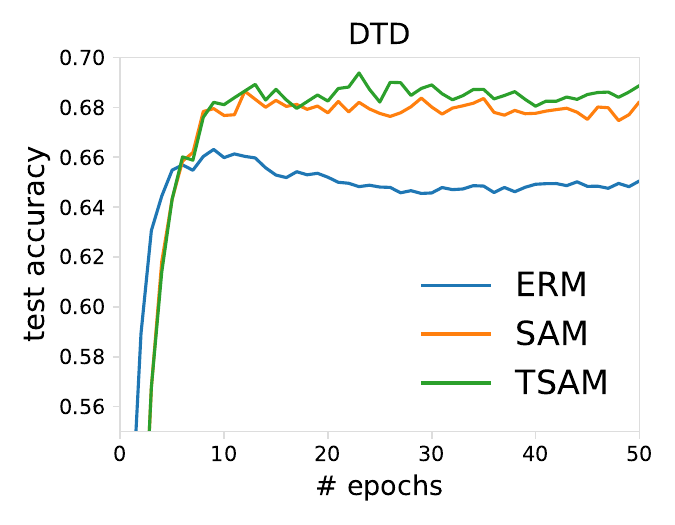}
    \end{subfigure}
    \hfill
    \begin{subfigure}{0.32\textwidth}
        \centering
        \includegraphics[width=\textwidth]{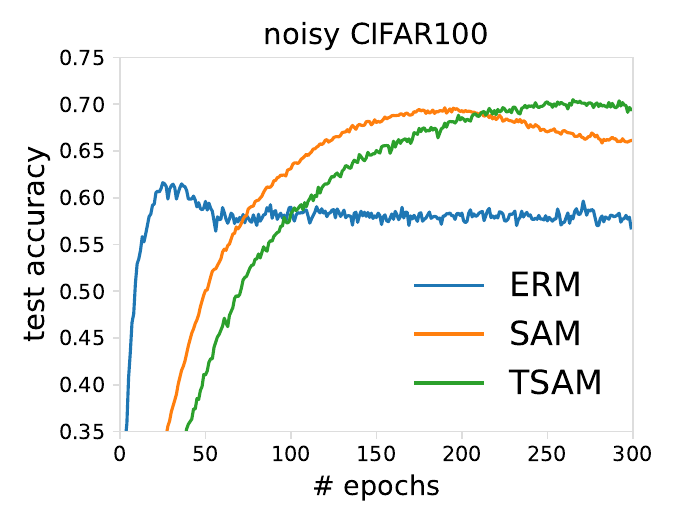}
    \end{subfigure}
    \caption{Convergence curves on three image datasets showing test accuracies.}
    \vspace{0.2in}
\label{fig:convergence}
\end{figure}

\paragraph{Convergence Curves.} In Table~\ref{table:main}, we present the final test accuracies of TSAM and the baseline. In Figure~\ref{fig:convergence}, we show the convergence of these methods on three datasets. We see that TSAM achieves the fastest convergence, and arrives at a better solution. This is consistent with our argument that TSAM with a bounded t is a more smooth objective than the original SAM formulation (Section~\ref{sec:properties}).

\newpage
\paragraph{Training and Test Loss Comparisons under Different Objectives.} In Table~\ref{table:train_test}, we show that TSAM generates better than ERM and SAM by comparing training and test losses.

\begin{table}[h!]
\caption{To further understand TSAM behavior, we report the losses of models trained by \{ERM, SAM, TSAM\} and evaluated on \{ERM, SAM, TSAM\} objectives, respectively. The left table shows \textbf{training} losses and the right one shows \textbf{test} losses. We see that (1) every objective achieves the smallest \textit{training} loss if directly being optimized (diagonal entries, left table). (2) Though SAM and TSAM incurs larger training losses than ERM (last two rows, left table), they lead to smaller test losses (last two rows, right table), i.e., better generalization. (3) Optimizing TSAM results in the smallest test loss across all the three metrics (last row, right table).}
\label{table:train_test}
    \centering
    \scalebox{0.99}{
    \begin{tabular}[t]{ c | c  c  c}
        \toprule[\heavyrulewidth]
         \multirow{2}{*}{\diagbox[innerwidth=3cm, height=2\line]{trained on}{evaluated on}} & ERM & SAM & TSAM  \\
         & \multicolumn{3}{c}{\textbf{training loss}} \\ \midrule
        ERM & \textbf{0.1283} & 1.35 &  3.48 \\
        SAM & 0.1489 & \textbf{0.22} &  0.60 \\
        TSAM & 0.1763  & 0.27 & \textbf{0.46} \\
        \bottomrule[\heavyrulewidth]
    \end{tabular}} \quad %
    \scalebox{0.99}{\begin{tabular}[t]{c | c  c  c}
         \toprule[\heavyrulewidth]
        \multirow{2}{*}{\diagbox[innerwidth=3cm,height=2\line]{trained on}{evaluated on}} & ERM & SAM & TSAM \\ 
        & \multicolumn{3}{c}{\textbf{test loss}} \\ \midrule
       ERM & 0.9302  & 2.05 &  3.54 \\
       SAM & 0.7414 & 0.91 & 1.34 \\
       TSAM & \textbf{0.7163} & \textbf{0.90} &  \textbf{1.08} \\
        \bottomrule[\heavyrulewidth]
    \end{tabular}}
\end{table}



{
\paragraph{Effects of $N$ in HMC.} Recall that Algorithm~\ref{alg:main} follows HMC updates in Eq. (\ref{eq:hmc_2}) by running it for $N=1$ step. In principle, we can run Eq. (\ref{eq:hmc_2}) for more than 1 step to generate each candidate perturbation $\epsilon$. This require additional gradient evaluations, which can be very expensive. We report results of $N=2$ and $N=3$ in Table~\ref{tab:N} below.
}
\begin{table}[h!]
    \centering
    \begin{tabular}{l | c c }
    \toprule[\heavyrulewidth]
       configurations  &   CIFAR100 & DTD \\
       \midrule
         $N=1~(s=3)$   & 0.7740 & 0.6882   \\ 
         $N=1~(s=5)$   & 0.7778 &  0.6870 \\ 
         $N=2~(s=3)$   & 0.7745 &  0.6883 \\  
         $N=3~(s=3)$   & 0.7752 &  0.6880 \\ 
    \bottomrule[\heavyrulewidth]
    \end{tabular}
    \caption{{$N>1$ requires $N$ times more gradient evaluations to generate a single perturbation $\epsilon$ (Eq. (\ref{eq:hmc_2}), without resulting in significantly better model accuracies.}}
    \label{tab:N}
\end{table}



\paragraph{Runtime Comparisons.}
We report the runtime of TSAM and other baselines in Table~\ref{table:runtime} below.

\begin{table}[!h]
    \caption{{Runtime comparisons (in minutes) averaged over different hyperparameter sets. ERM and SAM run the same numbers of epochs as TSAM. All baselines are explained in Section~\ref{sec:experiment}. ESAM1 denotes the baseline of letting SAM run longer until it reaches the same computation budget as TSAM. }
}
\centering
\label{table:runtime}
\scalebox{0.98}{
\begin{tabular}{lcccccccc}
	   \toprule[\heavyrulewidth]
        \textbf{datasets} & ERM & SAM & ESAM1 & ESAM2 & PGN & RSAM & TSAM \\
        \midrule
    CIFAR100  & 32  & 65   &  325  & 310 & 333 & 320 & 333 \\
  DTD  &  5 &  6 &  18 & 17 & 15 & 16 & 15 \\
  Noisy CIFAR100 & 45 & 92 & 350 & 315 & 302 & 312 & 337 \\
\bottomrule[\heavyrulewidth]
\end{tabular}}
\vspace{1em}
\end{table}

\begin{figure}[h!]
    \centering
    \begin{subfigure}{0.72\textwidth}
    \centering
        \includegraphics[width=\textwidth]{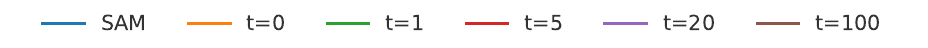}
    \end{subfigure}
    \centering
    \begin{subfigure}{0.33\textwidth}
        \centering
        \includegraphics[width=\textwidth]{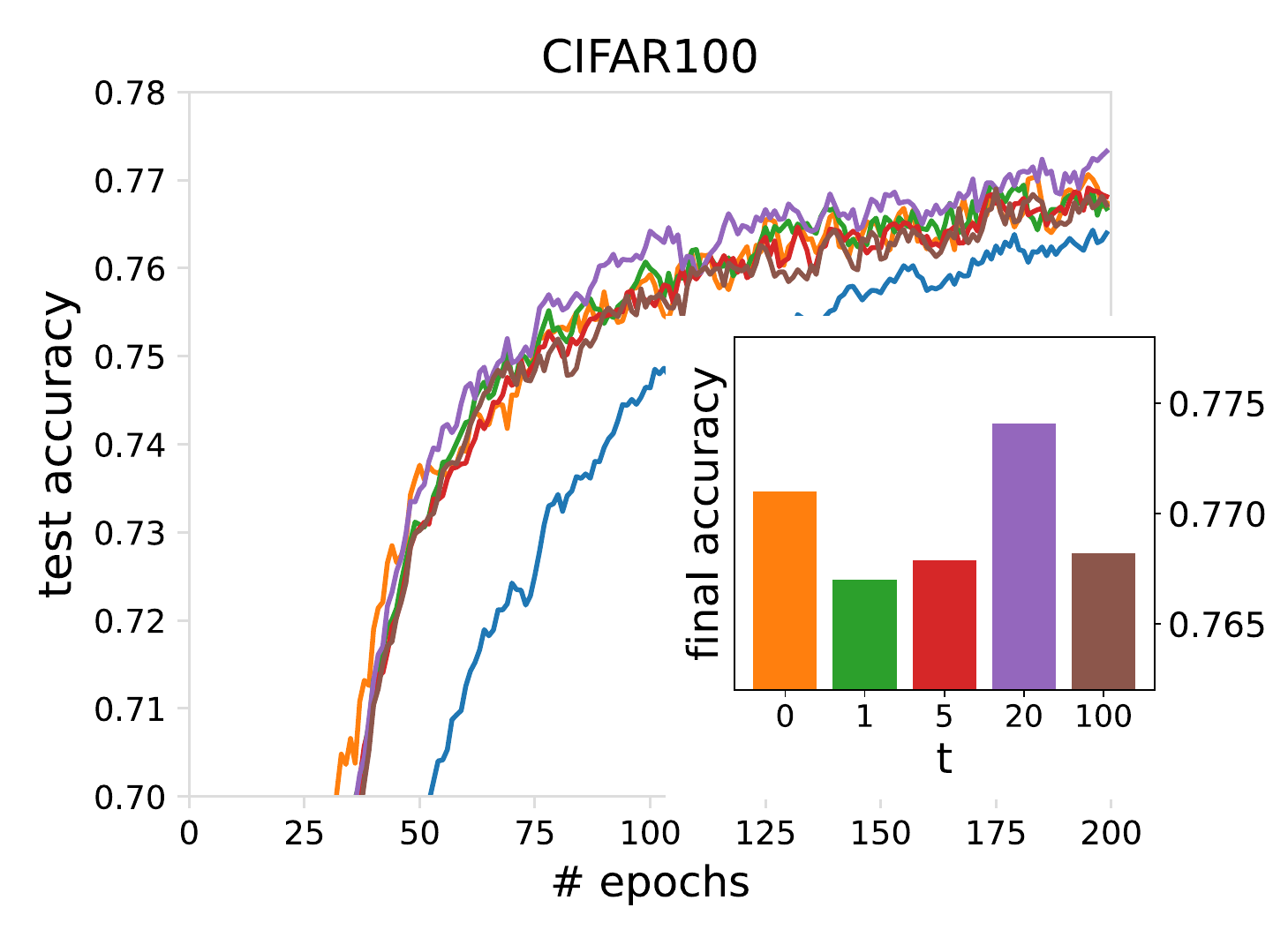}
    \end{subfigure}
    \begin{subfigure}{0.32\textwidth}
        \centering
        \includegraphics[width=\textwidth]{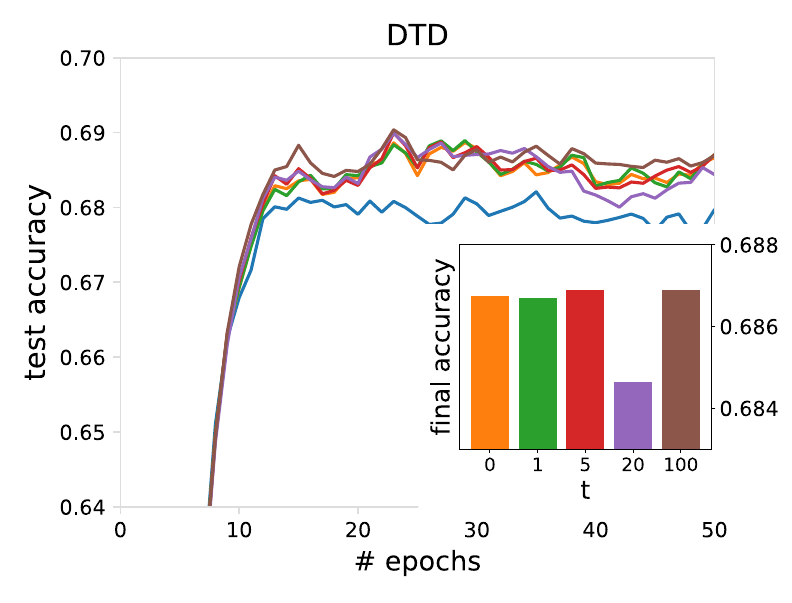}
    \end{subfigure}
    \begin{subfigure}{0.32\textwidth}
        \centering
        \includegraphics[width=\textwidth]{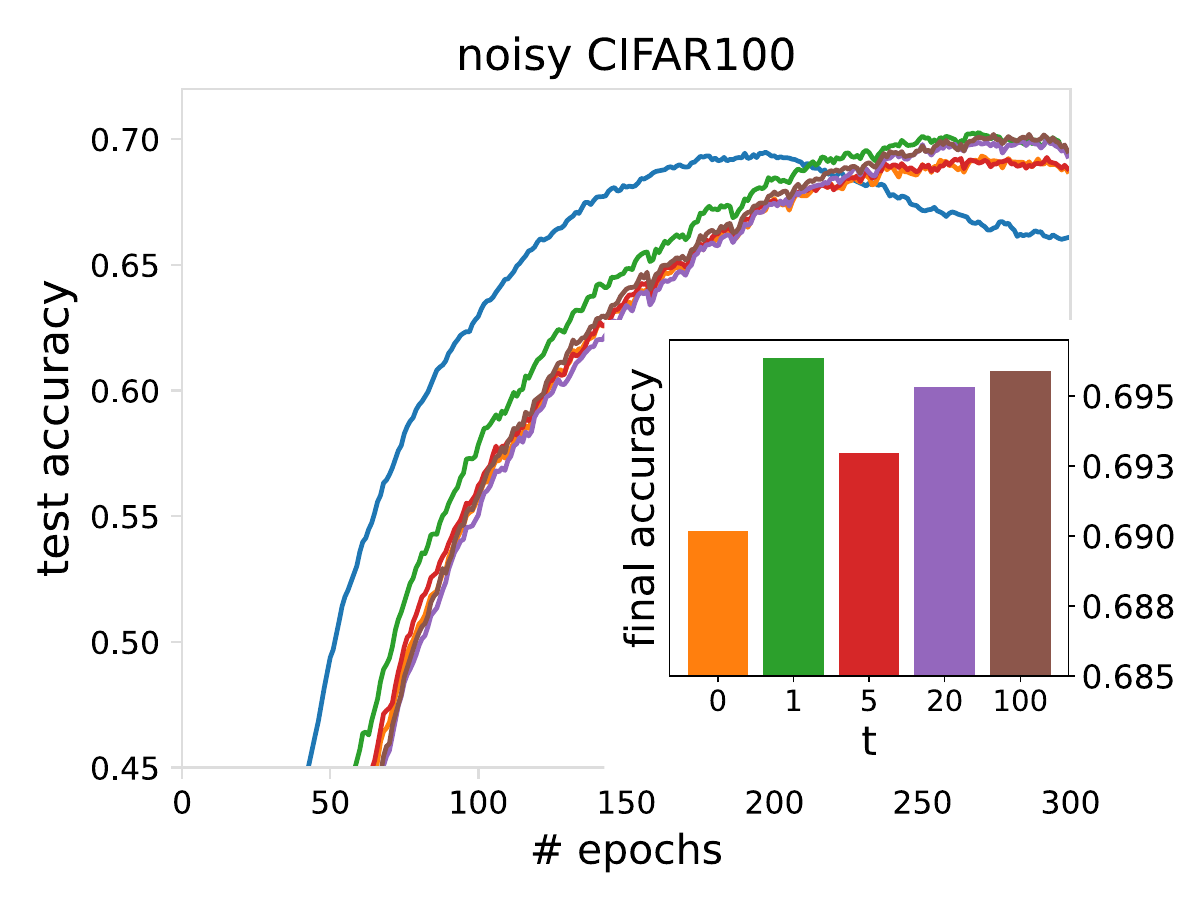}
    \end{subfigure}
    \caption{Test accuracies of SAM and TSAM for various values of $t$ when the number of sampled $\epsilon$'s is 3 for each dataset. We select the best SAM and TSAM runs based on the final accuracies on validation data. The results suggest that (1) there are multiple $t$ values that give superior performance than SAM; and (2) we typically need to manually tune a best $t$ via grid search. The empirical performance under different values of $t$ relies on tradeoffs between optimization efficiency (Section~\ref{sec:properties:general}, Section~\ref{sec:method}), flatness (Section~\ref{sec:properties:glms}), and generalization (Section~\ref{sec:properties:generalization}), and it is difficult to determine an optimal $t$ prior to training.
    Note that in the label noise regime (left subfigure), one might think that SAM performance could be further improved via a smaller learning rate or early stopping; however, we observe that SAM with a smaller learning rate does not give better accuracy. With early stopping, SAM accuracy is 0.6918, which is still 0.4\% lower than that of TSAM without early stopping. 
    }
\label{fig:multiple_t}
\end{figure}

\paragraph{Effects of the Tilting Hyperparameter $t$.} One critical hyperparameter in TSAM is $t$. When $t=0$, TSAM objective reduces to {the average-case perturbed objective}. When $t \to \infty$, the TSAM objective (Eq.~(\ref{obj:tsam})) recovers SAM (Eq.~\ref{obj:sam}). Here, we report the test accuracies as the training proceeds under multiple values of $t$'s for  the three tasks. In Figure~\ref{fig:multiple_t}, we see that there are a range of $t$'s that result in faster convergence or higher accuracies than SAM.

\begin{figure}[h!]
    \centering
    \includegraphics[width=0.35\linewidth]{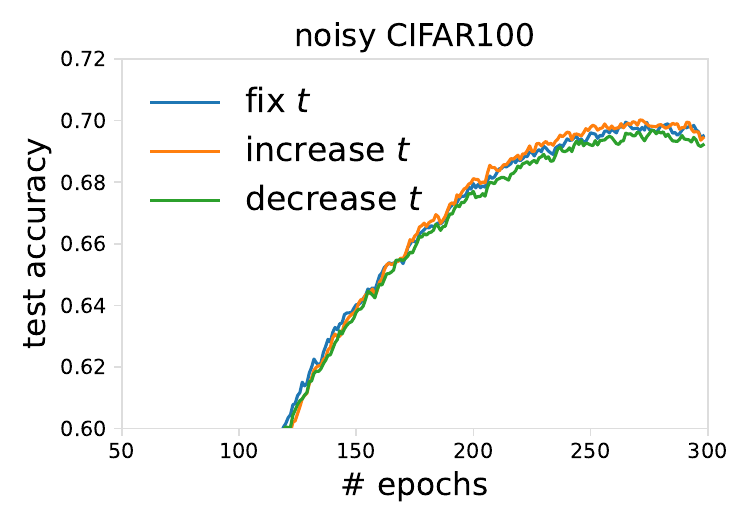}
  \vspace{-0.1in}
  \captionof{figure}{Linearly decreasing or increasing the tilting hyperparameter $t$ with the epochs  does not differ from the results of a fixed $t$.}
  \label{fig:schedult_t}
\end{figure}

\paragraph{Effects of Scheduling $t$.} We also evaluate the performance when we schedule $t$ dynamically during training. In Figure~\ref{fig:schedult_t}, we see that there is no significant difference between using a fixed $t$, increasing $t$ linearly, or decreasing it lienarly on the noisy CIFAR100 data.

\end{document}